%% file: main.tex
\input{generator/data}
\documentclass{article}
\usepackage{arxiv}

\usepackage{tikz}
\usetikzlibrary{shapes,arrows,calc}
\usetikzlibrary{positioning}
\usepackage{pgfplots}
\usepgfplotslibrary{colorbrewer}
\usetikzlibrary{pgfplots.statistics,pgfplots.colorbrewer} 
\usepackage{pgfplotstable}
\usepackage{filecontents}
\pgfplotsset{compat = 1.15, cycle list/Set1-8} 

\usepackage{graphicx,import,color}
\usepackage{xcolor}
\definecolor{codegreen}{rgb}{0,0.6,0}
\definecolor{codegray}{rgb}{0.5,0.5,0.5}
\definecolor{codepurple}{rgb}{0.58,0,0.82}
\definecolor{backcolour}{rgb}{0.95,0.95,0.92}

\usepackage{algorithm}
\usepackage[noend]{algpseudocode}
\usepackage{listings}
\usepackage{verbatim}

\usepackage{ulem}

\usepackage{adjustbox}
\usepackage{enumitem}
\usepackage{array}
\usepackage{subcaption}
\usepackage{csvsimple}
\usepackage{rotating}
\usepackage{booktabs} 
\usepackage{tablefootnote}

\usepackage{amsmath,amsthm,amssymb}
\usepackage{bm}
\usepackage{dsfont}

\newtheorem{definition}{Definition}[section]

\usepackage{hyperref}
\hypersetup{
    colorlinks=false,
    pdfborder={0 0 0},
    pdfborderstyle={/S/U/W 0},
}
\RequirePackage{doi}

\lstdefinestyle{mystyle}{
    backgroundcolor=\color{backcolour},   
    commentstyle=\color{codegreen},
    keywordstyle=\color{magenta},
    numberstyle=\tiny\color{codegray},
    stringstyle=\color{codepurple},
    basicstyle=\ttfamily\footnotesize,
    breakatwhitespace=false,         
    breaklines=true,                 
    captionpos=b,                    
    keepspaces=true,                 
    numbers=left,                    
    numbersep=5pt,                  
    showspaces=false,                
    showstringspaces=false,
    showtabs=false,                  
    tabsize=2
}
\newcount\Comments  
\newcommand{\kibitz}[2]{\ifnum\Comments=0\textcolor{#1}{#2}\fi}

\newcommand{\Hins}[1]{\kibitz{red}{[Hins: #1]}}

\input{macro}

\begin{document}

\title{Ride-pool Assignment Algorithms: Modern Implementation and Swapping Heuristics}

\author{
    Matthew Zalesak\\
    Operations Research and Information Engineering\\
    Cornell University \\
    Ithaca, NY, USA \\
    mdz32@cornell.edu\\
    \And
  Hins Hu\\
  Systems Engineering \\
  Cornell University \\
  Ithaca, NY, USA\\
  \texttt{zh223@cornell.edu} \\
   \And
  Samitha Samaranayake \\
  Civil and Environmental Engineering \\
  Cornell University \\
  Ithaca, NY, USA\\
  \texttt{samitha@cornell.edu} \\
}

\maketitle

\begin{abstract}
On-demand ride-pooling has emerged as a popular urban transportation solution, addressing the efficiency limitations of traditional ride-hailing services by grouping multiple riding requests with spatiotemporal proximity into a single vehicle. Although numerous algorithms have been developed for the Ride-pool Assignment Problem (RAP) -- a core component of ride-pooling systems, there is a lack of open-source implementations, making it difficult to benchmark these algorithms on a common dataset and objective. In this paper, we present the implementation details of a ride-pool simulator that encompasses several key ride-pool assignment algorithms, along with associated components such as vehicle routing and rebalancing. We also open-source a highly optimized and modular C++ codebase, designed to facilitate the extension of new algorithms and features. Additionally, we introduce a family of \textit{swapping-based local-search heuristics} to enhance existing ride-pool assignment algorithms, achieving a better balance between performance and computational efficiency.  Extensive experiments on a large-scale, real-world dataset from Manhattan, NYC reveal that while all selected algorithms perform comparably, the newly proposed \textit{Multi-Round Linear Assignment with Cyclic Exchange (LA-MR-CE)} algorithm achieves a state-of-the-art service rate with significantly reduced computational time. Furthermore, an in-depth analysis suggests that a performance barrier exists for all myopic ride-pool assignment algorithms due to the system's capacity bottleneck, and incorporating future information could be key to overcoming this limitation.

\end{abstract}

\section{Introduction}

On-demand ride-pooling services are an increasingly popular form of urban transportation, designed to enhance the efficiency of traditional ride-hailing systems by grouping passengers with similar routes into the same vehicle. From an operational standpoint, ride-pooling systems can vary in their level of central control. Some platforms function as two-sided marketplaces, connecting passengers with drivers, while others manage a centralized vehicle fleet, providing passengers with an interface for interaction. Regardless of the operational model, operators face a common challenge: making sequential decisions about assigning vehicles to passengers. This problem, commonly referred to as the \gls{a:rap}, has been extensively studied in both academic and industry contexts within the ride-sharing community.

In this paper, we consider the fully centralized scenario where the \gls{a:rap} admits the setting as follows: An operator manages a fleet of vehicles, and on-demand requests arrive in an online manner, each characterized by an origin, a destination, and a set of \gls{a:qos} constraints. The spatiotemporal distribution of requests is unknown to the operator in advance. The \gls{a:qos} constraints may include but are not limited to maximum waiting time, maximum detour time, and the number of passengers. The operator must respond to requests in (nearly) real-time, either accepting them with a guarantee that their \gls{a:qos} constraints are satisfied or rejecting them. In the backend, the operator has to determine an effective policy for request-vehicle assignment and vehicle routing, with the objective of optimizing one or more performance metrics, such as the daily service rate or the total vehicle miles traveled, depending on the operator's specific goals.

In most studies, such as \cite{alonso2017demand} \cite{riley2019column}, and \cite{shah2020neural}, the \gls{a:rap} is modeled as a \textit{multi-stage} optimization problem, where requests are batched over short periods, called \textit{decision epochs}, to enhance vehicle utilization and improve ride-sharing. A time-decomposed epoch-wise optimization problem is then solved dynamically. This scheme will be discussed in detail in Section \ref{sec:problem}. Despite the various models and algorithmic variants tackling the epoch-wise problem differently, none provide a comprehensive discussion of the implementation details or a well-organized open-source framework for practitioners. As a result, while these approaches share similar objectives, high-level strategies, and common components like vehicle simulation, benchmarking the performance of these algorithms within the same ride-pooling system using the same request data remains challenging. To address this gap, we present a modern implementation of a ride-pool assignment simulator, which includes a collection of commonly used algorithms and all the necessary components to operate a full-scale, real-world ride-pooling system. Our implementation, based on C++, is highly optimized for efficiency and modularized to facilitate extensions, enabling the incorporation of new state-of-the-art algorithms and features.

Furthermore, inspired by works such as \cite{simonetto2019real} and \cite{riley2019column}, which extended the \gls{a:rtv} algorithm from \cite{alonso2017demand} and balanced computational time with solution quality, we propose a family of ride-pool assignment algorithms based on swapping heuristics. We hypothesize that effectively swapping initial request-vehicle assignments, a local search strategy that has proven successful in various other domains, can achieve a better trade-off between computation time and solution quality in this context -- a balance that has been unexplored in previous studies. Our experiments confirm that certain swapping-based algorithms can match the solution quality of enumeration-based approaches like the full \gls{a:rtv}, while significantly reducing computational time.

The contributions of our paper to the ride-sharing literature are three-fold: 
\begin{enumerate}
    \item \textbf{Open-source Simulator:} We provide an open-source modern implementation of a ride-pool assignment simulator that facilitates the benchmarking of major algorithms.
    \item \textbf{Swapping-based Heuristics:} We propose a family of efficient swapping-based heuristics, expanding the repertoire of ride-pool assignment algorithms.
    \item \textbf{Comprehensive Experiments and Analysis:} We conduct extensive experiments on a real-world ride-pooling system in NYC under varying demand levels, evaluating the performance of all major algorithms and providing an in-depth quantitative analysis of their similarities and differences. This analysis offers valuable insights for guiding practical applications and future research directions.
\end{enumerate}

The paper is organized as follows: Section \ref{sec:literature} reviews related works in ride-pool assignment algorithms. Section \ref{sec:problem} provides a rigorous description of the \gls{a:rap}. Section \ref{sec:algos} reviews several existing popular ride-pool assignment algorithms, detailing their technical aspects. Section \ref{sec:swap} introduces a family of swapping heuristics first proposed in this paper. Section \ref{sec:stability} discusses the solution stability of the underlying \gls{a:ctsp} oracle and its impact on the effectiveness of the swapping heuristics and the multi-stage optimization scheme. Section \ref{sec:codebase} outlines the high-level structure of the open-source codebase. Section \ref{sec:experiments} presents the experimental results and subsequent analysis. Section \ref{sec:conclusion} concludes the paper and suggests future directions.

\section{Literature Review} \label{sec:literature}

In the past decade, numerous papers have been published to address the \gls{a:rap} and related problems, such as vehicle rebalancing, within the context of ride-sharing. To maintain focus and brevity, this review excludes traditional works on non-centralized taxi systems that predate the sharing economy era. The first scalable solution for real-world ride-pooling systems was proposed by \cite{ota2016stars}, which processes requests sequentially based on their emergence time and employs a greedy search for the best match. The seminal work in this field, \cite{alonso2017demand}, introduced a multi-stage optimization framework that integrates both request-vehicle assignment and idle vehicle rebalancing, setting a new foundation for subsequent research.

Since then, many works stemmed from \cite{alonso2017demand} aimed at either alleviating the computational burden (e.g., \cite{simonetto2019real}, \cite{riley2019column}, \cite{lowalekar2021zone}) or considering the non-myopic setting by incorporating future information (e.g., \cite{alonso2017predictive}, \cite{lowalekar2021zone}, \cite{yuen2019beyond}, \cite{shah2020neural}). \cite{simonetto2019real} employed a similar request-assignment pattern but reduced the problem space from a full \gls{a:rtv} graph to linear assignment. \cite{riley2019column} restructured the problem into a column generation framework and introduced a heuristic to generate high-quality trips for selection. \cite{lowalekar2019zac} utilized a zoning heuristic to simplify path construction and selection, though their solutions do not always guarantee compliance with the \gls{a:qos} constraints. \cite{alonso2017predictive} leveraged historical data to fit a distribution of future demand and integrated a set of predicted virtual near-future requests into the optimization. \cite{lowalekar2021zone}, as a follow-up to \cite{lowalekar2019zac}, extended the zone path model to the non-myopic setting. Instead of relying on shortest paths, \cite{yuen2019beyond} biased vehicle routes towards those expected to facilitate future demand matching. \cite{shah2020neural} modeled the problem using neural approximate dynamic programming, training a neural network surrogate to predict the decomposed value function of a vehicle. 


Additionally, we briefly review the works focusing on the underlying \gls{a:ctsp} related to ordering pick-up and drop-off locations. \cite{santi2014quantifying} introduced a heuristic based on a shareability graph, using a spatiotemporal index to efficiently identify the best vehicle for a request. \cite{tong2018unified} explored search procedures for pathfinding using insertion heuristics, which, while not optimal, are practical and feature low complexity. \cite{shen2019roo} concentrated on constructing valid ride-pooling paths through merge operations, beginning with the identification of shortest paths for each request and then building trip sets through sequences of merges.

\section{The Ride-pool Assignment Problem} \label{sec:problem}

We are given a road network $G = (N, A)$ with cost metric $c: N^2 \rightarrow \mathbb{R}_+$ representing the shortest-path travel time between two nodes, an operational time horizon $[0, T]$, and a collection of centrally controlled vehicles $\mathcal{V}$ where each one associated with a capacity $Q_v$. Each vehicle $v \in \mathcal{V}$ has an initial position $n^0_v \in N$ and is assumed to be empty at time $t = 0$.

Over the horizon, there is a set of requests $\mathcal{R}$ streamed to the operator in an online manner. The underlying spatial-temporal distribution $\mathcal{D}_{\mathcal{R}}$ is unknown to the operator at time $t = 0$. Each request $r = (o_r, d_r, t_r, \mathcal{C}_{QoS})$  is characterized by an origin $o_r \in N$, a destination $d_R \in N$, the emergence time $t_r$ when it becomes visible to the operator, and a set of \gls{a:qos} constraints specified by passengers. For simplicity, we only treat $\mathcal{C}_{QoS}$ as including only two crucial constraints in this paper: the maximum waiting time $\Delta_r^w$ and the maximum detour time $\Delta_r^d$. Denote by $l_r^b$, $l_r^a$, and $\bar{t}_r$ the latest boarding time, the latest arrival time, and the actual boarding time of request $r$ respectively. In a deterministic system, the following relationships can be observed:
\begin{equation} \label{eq:max_wait_detour}
    l_r^b = t_r + \Delta_r^w \quad l_r^a = \bar{t}_r + c(o_r, d_r) + \Delta_r^d
\end{equation}
As aforementioned, the \gls{a:rap} is often framed as a multi-stage optimization problem. In this setting, the entire horizon $[0, T]$ is partitioned into a set of uniformly spaced decision epochs $E$. Non-uniform and adaptive partitioning can also be presumed when extraneous information is available to improve decision-making, although this is beyond the scope of this paper. Let $[t_e^1, t_e^2]$ represent the start time and the end time of epoch $e \in E$, respectively. The length of each epoch, $t_e^2 - t_e^1$, referred to as the \textit{request batching interval}, is determined in practice based on the system's scale. At the end time of each epoch, the operator can view all requests that have become visible up to that time, but can only possibly serve those that persist beyond the epoch. Formally, the subset of emerging requests during epoch $e$ is defined as $\mathcal{R}_e = \{r \in \mathcal{R} \mid (t_e^1 < t_r \leq t_e^2) \wedge (t_e^2 \leq l_r^b)\}$. The second condition filters out requests that disappear too quickly. In other words, we assume the batching interval is typically shorter than the persistence duration of any request.

Then, the operator solves an \textit{epoch-wise \gls{a:rap}} at each epoch to simultaneously assign available vehicles to requests and to determine vehicle routes to serve the requests. In the solution, a request can be \textit{unserved} if no vehicle in the fleet can feasibly serve it, and a vehicle can also be left \textit{unassigned} if supply exceeds demand either locally or globally, depending on the \gls{a:qos} constraints. It is important to explicitly note that a vehicle may have received assignments in previous epochs, even if it remains unassigned in the current epoch. Unserved requests are carried over to the next epoch $e^{'}$ if their latest boarding times exceed the end time of $e^{'}$, except for the case where a full \gls{a:rtv} framework is adopted \footnote{In the full \gls{a:rtv} framework \cite{alonso2017demand}, we can compute the future states of all vehicles under every possible assignment. If we cannot serve a request in one decision epoch, no new assignment possibility will appear in future epochs, and thus the request will still not be served. More details will be demonstrated in Section \ref{sec:rtv}.}. Unassigned vehicles either continue their route if they have previous assignments or move to other areas for \textit{rebalancing}. The rebalancing component is used to relocate idle vehicles to other high-demand areas to potentially yield a better request-vehicle assignment in future epochs, which will be discussed in detail in Section \ref{sec:rebalancing}.

Typically, effective algorithms address the epoch-wise \gls{a:rap} by decomposition, breaking down the decision problem into request-vehicle assignment and vehicle routing. An assignment of a vehicle $v$ to a subset of new requests $\mathcal{R}^{'} \subseteq \mathcal{R}_e$ is called a \textit{trip}, defined as $f = (v, \mathcal{R}^{'})$. Meanwhile, the feasibility and the cost of routing a vehicle to serve those new requests in $\mathcal{R}^{'}$ depend on the state of the vehicle, including the current location and onboard requests, which can be pre-determined by solving a \gls{a:ctsp} sub-problem. In particular, the routing instructions provided to vehicle $v$ by the solution to a \gls{a:ctsp} is called a \textit{vehicle route}, defined as a sequence of stops $S_v = (n_1, \dots, n_k)$, where $n_i \in N$, $i \in [k]$, and $k \leq Q_v$. Each stop $n_i$ includes detailed information about the corresponding request, such as the action (i.e., pick-up or drop-off) and the emergence time. Given the current route $S_v$ and the updated route $S^{'}_v$ of adding a trip $f$, the trip cost is defined as $c_f = c(S^{'}_v) - c(S_v)$, where $c(S) = \sum_{n_i \in S} c(n_i, n_{i+1})$.

Overall, the objective of the \gls{a:rap} is to optimize one or more performance metrics that vary across different ride-pooling systems. One widely accepted option is a multi-level objective that maximizes the total number of served requests over the planning horizon, namely the \gls{a:sr}, while minimizing the total \gls{a:vmt} \cite{alonso2017demand}. The maximization (i.e., primary objective) is prioritized over the minimization (i.e., secondary objective) as a system with a higher \gls{a:sr} is considered more effective despite potentially higher costs.


\section{Existing Ride-pool Assignment Algorithms} \label{sec:algos}

With the problem setting described above, we review several state-of-the-art algorithms designed to address the request-vehicle assignment component in this section. For comprehensiveness, we also include a discussion of the vehicle rebalancing algorithm as a special case of assignment, which is detailed in Section \ref{sec:rebalancing}. The vehicle routing component is discussed separately in Section \ref{sec:stability}. These algorithms distinguish themselves in their approaches to addressing the epoch-wise \gls{a:rap} while maintaining consistency within the multi-stage framework. We will focus more on the implementation while limiting our discussion of other aspects, such as the underlying mathematics and analysis, to a concise and self-consistent scope. First, we review the full \gls{a:rtv} algorithm \cite{alonso2017demand} that solves the epoch-wise problem exactly. As the exact formulation may become intractable as the system scales, we identify some heuristics, including those used in \cite{alonso2017demand}, that can be used in conjunction with the algorithm. However, it can be a difficult engineering practice to design these heuristics in order to strike the right balance between achieving a better epoch-wise objective and saving more computation time. Thus, we further cover two non-exact algorithms, based on \gls{a:cg} \cite{riley2019column} and \gls{a:la} \cite{simonetto2019real}, respectively, which implicitly consider the trade-off between solution quality and efficiency. Both of them are extended from \cite{alonso2017demand} but are generally recognized as frameworks in their own right.

\subsection{The \gls{a:rtv} Algorithm} \label{sec:rtv}
The full \gls{a:rtv} algorithm is the pioneering algorithm that provides anytime-optimality guarantees by exploring all possible trips and assignments when no timeout is set. In each decision epoch $e \in E$, the algorithm is executed through the following pipeline:
\begin{enumerate}[label={(\arabic*)}]
    \item \textbf{Identify Unpicked-up Requests:} Determine the set of requests $\overline{\mathcal{R}}$ that are assigned in previous epochs but have not been picked up yet. 
    
    \item \textbf{Construct Shareability Graph:}  Create a sharability graph on the node set $\overline{\mathcal{R}} \cup \mathcal{R}_e$. An edge is added between pairs of requests if the two requests $r$ and $r^{'}$ could feasibly share a trip in a virtual empty vehicle $v$ located at the origin of either request (i.e., either $o_r$ or $o_{r^{'}}$). The notion of shareability graph was introduced in \cite{santi2014quantifying}

    \item \textbf{Extend Shareability Graph:} Add request-vehicle edges to the graph. An edge is added between a vehicle $v$ and a request $r$ if there exists a feasible trip routing $v$ to serve $r$. Note that passengers already in the vehicle $v$ are considered implicitly.

    \item \textbf{Enumerate Candidate Trips:} Initialize an empty set $\mathcal{F}_e$ of candidate trips. Enumerate all cliques \footnote{Mathematically, a clique in a graph $G$ is a complete subgraph $C \subseteq G$.} in the shareability graph consisting of exactly one vehicle and one or multiple requests. If a (shared) trip $f$ represented by a clique is feasible, we add it to the set $\mathcal{F}_e$ and record the trip cost $c_f$.

    \item \textbf{Construct RTV Graph:} Create the \gls{a:rtv} graph by combining the request set $\overline{\mathcal{R}} \cup \mathcal{R}_e$, the vehicle set $\mathcal{V}$, and the set $\mathcal{F}_e$ of candidate trips. Add an edge between $r \in \overline{\mathcal{R}} \cup \mathcal{R}_e$ and $f \in \mathcal{F}_e$ if request $r$ is served in trip $f$. Add an edge between $v \in \mathcal{V}$ and $f \in \mathcal{F}_e$ if trip $f$ involves vehicle $v$.
    
    \item \textbf{Solve ILP Model:} Given the \gls{a:rtv} graph, solve the \gls{a:ilp} (\ref{eq:ilp}) to obtain the optimal epoch-wise ride-pool assignment. 
\end{enumerate}
Including the request set $\overline{\mathcal{R}}$ enables \textit{implicit re-assignment}. In principle, the assignment of a request can be changed until its latest boarding time approaches. While this characteristic might be less suitable for two-sided marketplaces, it is advantageous in a fully centralized system. This flexibility allows the system to make smarter decisions as new information becomes available, though it may result in some passengers being rescheduled multiple times within their tolerance limits.

In practice, every query of a trip, including trip generation, cost computation, and trip feasibility checking, involves calling an oracle that solves the underlying \gls{a:ctsp}. Hence, the enumeration in step (3) can proceed concurrently with the shareability graph construction in steps (1) and (2), avoiding redundant calls to the \gls{a:ctsp} solver.

The algorithm's exactness arises from enumerating all feasible trips involving any number of requests within the current epoch in step (3). However, in contrast to a naive enumeration, step (3) limits the search to cliques in the shareability graph, generating candidate trips in a constrained space, such that no trip consisting of $k$ requests is considered feasible if any subset consisting of $k^{'} < k$ requests cannot share a trip. Indeed, this is the recipe that makes the algorithm tractable for reasonably sized ride-pooling systems, as the number of candidate trips is limited by the tight service time windows of on-demand requests. 

\subsubsection{The \gls{a:ilp} of Epoch-Wise Ride-pool Assignment}
Define $x_f$ as a binary variable for each trip $f \in \mathcal{F}_e $, where $x_f = 1$ indicates that the trip is selected. Define $y_r$ as a binary variable for each $r \in \mathcal{R}_e$, where $y_r = 1$ indicates that request $r$ is unserved in epoch $e$. Denote by $M$ a substantial penalty coefficient for each unserved request.  Denote by $c_f$ the cost of routing vehicle $v$ to serve trip $f$. The minimum cost epoch-wise ride-pool assignment can be formulated by the \gls{a:ilp} shown below.

\begin{equation} \label{eq:ilp}
\renewcommand{\arraystretch}{2}
\begin{array}{rcc}
    \min\limits_{\bm{x}} & \sum\limits_{f \in \mathcal{F}_e} c_f x_f  + M \sum\limits_{r \in \mathcal{R}_e} y_r \\
    \text{s.t.} 
    & \sum\limits_{f \in \mathcal{F}_e \, : \, v \in f} x_f \leq 1 & \forall \; v \in \mathcal{V} \\
    & \sum\limits_{f \in \mathcal{F}_e\, : \, r \in f} x_f + y_r = 1 & \forall \; r \in \mathcal{R}_e \\
    & x_f \in \{0, 1\}^{|\mathcal{F}_e|}, \quad y_r \in \{0, 1\}^{|\mathcal{R}_e|}
\end{array}
\end{equation}
The presence of a substantial penalty term, along with the second set of constraints, ensures the optimal solutions prioritize the maximization of service rate over the minimization of the trip cost. Although this \gls{a:ilp} is NP-hard since it generalizes the Weighted Set Packing Problem, many modern \gls{a:mip} solvers can efficiently find (near-)optimal solutions to real-world size instances in practice. 

\subsubsection{Practical Heuristics for the RTV Algorithm}

Despite the sparsity of the shareability graph in many real-world ride-pooling systems, applying the full \gls{a:rtv} algorithm to obtain an exact solution remains impractical for large-scale systems, such as a centralized ride-sharing service in the Manhattan area. In this case, heuristics are required to further prune the set of candidate trips and reduce the computational burden. We briefly outline several such heuristics below.
\begin{enumerate}[label={(H\arabic*)}]
    \item For every oracle call regarding a trip, an exact \gls{a:ctsp} solver can be replaced by a heuristic solver.
    \item In steps (1) and (2) in the aforementioned pipeline, we can construct an approximate shareability graph instead of an exact one.
    \item In step (3) in the aforementioned pipeline, we can enforce a computational time limit for candidate trip generation.
    \item Within the time limit set in step 3, we can devise an ordering heuristic to increase the likelihood of efficient trips being generated first.
    \item Do not allow the implicit re-assignment even though a request $r \in \overline{\mathcal{R}}$ assigned to a trip during a previous epoch has not been picked up yet.
\end{enumerate}

(H1), (H2), and (H3) are all mentioned in \cite{alonso2017demand}. We will provide additional implementation details for (H1) in Section \ref{sec:stability}. (H2) is not covered in this paper but is proven to be effective in an \gls{a:rtv}-based algorithm for the school bus routing problem, as shown in \cite{guo2018solving}. (H3) is called \textit{Fast RTV} and will be benchmarked against the full version in Section \ref{sec:experiments}. (H4) is broadly defined and can be implemented in various ways. Among all possibilities, \cite{riley2019column} introduces a \gls{a:cg} framework that aligns with the core idea of this heuristic. Instead of generating a static set of candidate trips in step (3) and feeding the entire set to the \gls{a:ilp} in step (4), one can dynamically generate efficient trips through a \textit{pricing} sub-routine and solve restricted \gls{a:ilp}s in rounds. The rationale is that many low-value candidate trips, even though feasible, do not contribute to the optimal solution to \gls{a:ilp} (\ref{eq:ilp}). Due to its sophistication, this extended framework is regarded as an independent non-exact algorithm. We will cover more details in Section \ref{sec:cg}. (H5) can be directly incorporated into the \gls{a:rtv} algorithm, but its more significant application is within the \gls{a:la} framework introduced by  \cite{simonetto2019real}, which simplifies \gls{a:rtv}. Similar to the \gls{a:cg} framework, \gls{a:la} is also considered as an independent non-exact algorithm in the literature. We will discuss it in detail in Section \ref{sec:la}. 

\subsection{Column Generation} \label{sec:cg}
Column generation can date back to the 1950s when it was originally designed for large-scale \gls{a:lp}. Over time, it has evolved into an optimization framework capable of addressing large-scale \gls{a:ilp} \cite{barnhart1998branch} as well, namely Branch-and-Price. In general, the target integer optimization problem, assumed to be a minimization without loss of generality, needs to be transformed into a partition-based formulation where each decision variable represents a partial solution (i.e., a column). Typically, the number of such variables grows exponentially with problem size, and thus getting an exact solution is almost impossible. To manage this, \gls{a:cg} considers a \textit{\gls{a:rmp}} in the same form but consisting of only a subset of decision variables. Starting with any initial small subset that admits a feasible solution, the method iteratively adds one or more efficient columns to the \gls{a:rmp} and re-optimizes it using conventional branch-and-cut algorithms. The initial subset can be determined arbitrarily or through extra heuristics. In each iteration, one needs to solve an auxiliary optimization problem called the \textit{pricing problem} to determine the efficient columns to be added. Across all possible columns, the rigorous form of the pricing problem attempts to find the one with minimum negative \textit{reduced cost} with respect to the dual \gls{a:lp} relaxation of the \gls{a:rmp} at the same iteration. The algorithm continues until either a computational time limit is reached or the pricing problem no longer finds columns with negative reduced cost. When the latter occurs, the incumbent solution is guaranteed to be optimal for the original problem.

In the work by \cite{riley2019column}, they directly use the trip-based formulation as presented in \gls{a:ilp} (\ref{eq:ilp}) as \textbf{each trip itself is a column}. In the remainder of this section, the terms ``trip'' and ``column'' are used interchangeably. Assume $\mathcal{F}^{k}$ and $\mathcal{R}^{k}$ are the set of selected trips and the set of corresponding requests in the \gls{a:rmp} at iteration $k$ of \gls{a:cg}, the dual \gls{a:lp} relaxation is shown below in (\ref{eq:dual_lp}). $\pi_r$ are dual variables corresponding to the first constraint set in \gls{a:ilp} (\ref{eq:ilp}) and $\sigma_v$ are dual variables corresponding to the second constraint set in \gls{a:ilp} (\ref{eq:ilp}). The objective is exactly the reduced cost. The notation $v_f$ represents the vehicle assigned to the trip $f$.
\begin{equation} \label{eq:dual_lp}
\renewcommand{\arraystretch}{2}
    \begin{array}{rcc}
        \min\limits_{\bm{\pi}, \; \bm{\sigma}} & (c_f - \sum\limits_{r \in \mathcal{R}^k} \pi_r - \sigma_{v_f}) \\
        \text{s.t.} 
        & \pi_r \leq M & \forall \; r \in \mathcal{R}^k \\
        & \sum\limits_{r \in f} \pi_r + \sigma_{v_f} \leq c_f & \forall \; f \in \mathcal{F}^k \\
        & \sigma_v \leq 0 & \forall \; v \in f
    \end{array}
\end{equation}

Given the solution $\bm{\pi}^*$ and $\bm{\sigma}^*$ to \gls{a:lp} (\ref{eq:dual_lp}) at iteration $k$ of \gls{a:cg}, the pricing problem is to search for trips such that their reduced costs $(c_f - \sum_{r \in \mathcal{R}^k} \pi_r -\sigma_{v_f})$ are negative. The search space is $\mathcal{F}_e \setminus \mathcal{F}^k$ where $\mathcal{F}_e$ is defined in Section \ref{sec:rtv}. Note that the size of this set is approximately the same order as that of the entire set of candidate trips as we assume $\mathcal{F}^k$ to be relatively small in \gls{a:cg}. The problem is NP-hard because determining the trip cost $c_f$, which involves solving a \gls{a:ctsp}, is itself NP-hard. In fact, it is hypothesized that even approximating this problem is hard \cite{martinez2020request}. The NP-hardness of the pricing problem complicates the application of \gls{a:cg} in the epoch-wise \gls{a:rap}, as finding the optimal column or any column with a negative reduced cost may not be straightforward through directly solving the pricing problem.


Hence, \cite{riley2019column} proposes a heuristic, as a surrogate solver of the pricing problem, to generate efficient columns that respect their prices.  With this heuristic integrated, the entire \gls{a:cg} algorithm is shown in Algorithm \ref{alg:cg}. In each iteration $k$ of \gls{a:cg}, procedure \textsc{GenerateColumns} in lines 1 to 9 calls procedure \textsc{GenerateSizedColumns} to generate multiple candidate trips (i.e., columns) consisting of $j$ requests. This starts with $j = 1$ and increases $j$ by $1$ until $j$ equals the total number of requests, continuing only if no column is found for the current $j$. It returns the set of columns as soon as at least one column is found. The generated columns are then added to the \gls{a:lp} relaxation of the \gls{a:rmp} for re-optimization, and the \gls{a:cg} algorithm proceeds to the next iteration $k+1$. If, during any iteration $k^{'}$, procedure \textsc{GenerateColumns} fails to return any column or the computational time limit is reached, the \gls{a:cg} algorithm terminates. Due to the complexity arising from a strict branch-and-price, \cite{riley2019column} only applies \gls{a:cg} at the root node without branching involved. At the last step, the integrality constraint is added back, and the \gls{a:rmp} with the currently generated columns is solved to obtain a final solution.


Lines 16 to 24 describe \textsc{GenerateSizedColumns} in detail. Given a fixed number $j$, the procedure iteratively generates an efficient column for each vehicle, ensuring that (1) all columns consist of disjoint sets of requests, and (2) each column consists of no more than $j$ requests. Vehicles with larger corresponding dual variables $\sigma_v$ are prioritized to include requests in their columns. The procedure \textsc{CTSPSolver} in line 19 is the \gls{a:ctsp} oracle discussed in Section \ref{sec:problem}. It takes a vehicle and a request set as input and returns the (nearly) best configuration of a column. 

\begin{algorithm}[!htbp]
\caption{Column Generation}
\label{alg:cg}
\begin{algorithmic}[1]
\setlength{\baselineskip}{1.1\baselineskip} 
\State $k = 1$
\While{\textsc{TimeOut} = \text{False}}
\State $\mathcal{C} \gets \textsc{GenerateColumns}()$
\If{$\mathcal{C} = \varnothing$}
    \State \textbf{Return} $\bm{x}^* \gets \text{\gls{a:rmp}}(k)$
\EndIf
\State $k = k + 1$
\State Solve \gls{a:rmp}($k$) after adding $\mathcal{C}$
\EndWhile
\vspace{0.5em}
\Procedure{GenerateCoumns()}{}
    \State $j \gets 1$
    \While{$j \leq |\mathcal{R}_e|$}
        \State $\mathcal{C} \gets \textsc{GenerateSizedColumns}(j)$ 
        \If{$\mathcal{C} \neq \varnothing$}
            \State \textbf{Return} $\mathcal{C}$
        \Else
            \State $j = j + 1$
        \EndIf
    \EndWhile
\EndProcedure
\vspace{0.5em}
\Procedure{GenerateSizedColumns}{$j$}
    \State $\mathcal{C} \gets \varnothing, \quad \mathcal{R}(j) = \{S \subseteq \mathcal{R}_e \; | \; |S| = j\}$
    \For{$v \in \mathcal{V}$ ordered by decreasing $\sigma_v$}

        \State $\mathcal{F}(v) \gets \{f = \textsc{CTSPSolver}(v, S) \; | \; S \subseteq \mathcal{R}(j) \}$
        \State $f^* \gets \arg\min_{f \subset \mathcal{F}(v)} \textsc{ReducedCost}(f)$ 
        \If{$\textsc{ReducedCost}(f^*) < 0$}
            \State $\mathcal{C} \gets \mathcal{C} \cup \{f\}$
            \State $\mathcal{R}(j) \gets \mathcal{R}(j) \setminus \{r\; | \; r \in f^*\}$
        \EndIf
    \EndFor
    \State \textbf{Return} $\mathcal{C}$ 
\EndProcedure
\end{algorithmic}
\end{algorithm}

\subsection{Linear Assignment} \label{sec:la}

The linear assignment (LA) algorithm for the epoch-wise \gls{a:rap}, first demonstrated in \cite{simonetto2019real}, offers substantial computational time savings by bypassing the costly trip generation phase used in the full \gls{a:rtv} and \gls{a:cg} algorithms. Rather than generating shared trips involving multiple requests, \gls{a:la} only considers trips formed by adding a single request to an existing vehicle route in each decision epoch. Additionally, if any requests emerging in the previous epoch are not served and have not expired (i.e., the latest pick-up time is later than the start time of the current epoch), they are carried over to the current epoch. Consequently, the set of all candidate trips considered in \gls{a:la} is defined by (\ref{eq:la}), where $\hat{\mathcal{R}}$ is the set of all unserved requests emerging in the previous epochs and being carried over, and $(v, r)$ denotes adding request $r$ to the existing route of vehicle $v$. The size of $\mathcal{F}_e$ is significantly smaller than the counterpart in the full \gls{a:rtv} algorithm if the shareability graph is not sparse. However, given that sparsity is a common property in most shareability graphs in real-world ride-pooling systems, using \gls{a:la} is unlikely to significantly compromise the quality of the solution to the epoch-wise \gls{a:rap}. 
\begin{equation} \label{eq:la}
    \mathcal{F}_e = \{f = (v, r) \; | \; f \; \text{is feasible}, \; \forall \; v \in \mathcal{V}, \; \forall \; r \in \mathcal{R}_e \cup \hat{\mathcal{R}}\}
\end{equation}

We can then formulate another \gls{a:ilp} as shown in (\ref{eq:la_ilp}). Compared to the formulation in \gls{a:rtv}, this one uses pair subscripts to denote the trip cost $c_{vr}$ and the binary decision variable $x_{vr}$, highlighting the resemblance of the problem to bipartite matching. When a specific assignment $(v, r)$ is infeasible, the corresponding cost $c_{vr}$ is set to infinity. In practice, if the \gls{a:lp} solver does not support infinite values, these can be replaced with a sufficiently large number.
\begin{equation} \label{eq:la_ilp}
\renewcommand{\arraystretch}{2}
\begin{array}{rcc}
    \min\limits_{\bm{x}} & \sum\limits_{(v, r) \in \mathcal{F}_e} c_{vr} x_{vr} \\
    \text{s.t.} 
    & \sum\limits_{r: (v, r) \in \mathcal{F}_e} x_{vr} \leq 1 & \forall \; v \in \mathcal{V} \\
    & \sum\limits_{v: (v, r) \in \mathcal{F}_e\, : \, r \in f} x_{rv} \leq 1 & \forall \; r \in \mathcal{R}_e \cup \hat{\mathcal{R}} \\
    & x_{vr} \in \{0, 1\}^{|\mathcal{F}_e|}
\end{array}
\end{equation}

Although (\ref{eq:la_ilp}) is an \gls{a:ilp}, its \gls{a:lp} relaxation is tight. Essentially, the problem is a Maximum Weighted Bipartite Matching Problem. Vehicles and requests form two disjoint sets of nodes and trips are the edges of the bipartite graph. Many algorithms can solve it in polynomial time, provided they have access to the \gls{a:ctsp} oracle to compute the edge weights. For better code encapsulation and reusability, however, we solve it by feeding the \gls{a:ilp} into a \gls{a:mip} solver in our implementation, which has proven to be highly efficient in practice. Furthermore, for every request carried over from a previous epoch, we multiply the cost by a constant to prioritize them over the new requests. 

Unlike the full \gls{a:rtv} and the \gls{a:cg} algorithms, \gls{a:la} never reassigns requests to different vehicles in later epochs, even if they have not yet boarded. This approach is the key point making \gls{a:la} successful because implicit reassignment results in a buildup of old requests, negatively impacting system performance.

Despite its simplicity and speed-up, \gls{a:la} has some drawbacks. One key concern is that \gls{a:la} may impose an artificial constraint on the assignments, where a request that pairs well with others (i.e., people from the same neighborhood) must be assigned to a different vehicle, leading to a potentially suboptimal global solution. This issue can be avoided by full \gls{a:rtv}, \gls{a:cg}, or even a greedy algorithm. A greedy algorithm can be seen as a naive version of \gls{a:la}, which iteratively assigns the most cost-efficient vehicle to the first request in the stream without batching. Furthermore, using \gls{a:la} requires a careful design of the request batching interval to balance supply and demand. An interval that is too short may be overly greedy while an interval that is too long risks losing many requests with short waiting times. These limitations motivate us to devise swapping heuristics that find a middle ground between the full-scaled algorithms like full \gls{a:rtv} and \gls{a:cg} and the baseline algorithms like \gls{a:la}.

\subsection{Vehicle Rebalancing} \label{sec:rebalancing}
After the request-vehicle assignment in every epoch, fleet imbalances may result in both a subset of unassigned requests $\mathcal{R}_{un}$, and a subset of unassigned idle vehicles $\mathcal{V}_{un}$. This occurs when these idle vehicles are too far away from the regions where the unserved requests have just emerged. To potentially mitigate such imbalances in future epochs, \cite{alonso2017demand} proposed solving a vehicle rebalancing \gls{a:lp} to relocate idle vehicles to low-supply regions after the regular assignment. The formulation is presented in \gls{a:lp} (\ref{eq:rebalancing_lp}). In particular, $\tau_{vr}$ represents the shortest path travel time from the current location of vehicle $v$ to the origin of request $r$. The optimal solution to \gls{a:lp} (\ref{eq:rebalancing_lp}) contains integers only, where $x_{vr}$ indicates relocating vehicle $v$ to the origin of request $r$. There exist other vehicle rebalancing algorithms, such as \cite{JIAO2021103289}, that utilize more sophisticated frameworks. They are beyond the scope of this paper.
\begin{equation} \label{eq:rebalancing_lp}
\renewcommand{\arraystretch}{2}
\begin{array}{rc}
    \min\limits_{\bm{x}} & \sum\limits_{v \in \mathcal{V}_{un}} \sum\limits_{r \in \mathcal{R}_{un}} \tau_{vr} x_{vr} \\
    \text{s.t.}
    &  \sum\limits_{v \in \mathcal{V}_{un}} \sum\limits_{r \in \mathcal{R}_{un}} x_{vr} = \min(|\mathcal{R}_{un}|, |\mathcal{V}_{un}|) \\
    & 0 \leq x_{vr} \leq  1
\end{array}
\end{equation}

\section{The Swapping Heuristics} \label{sec:swap}
In this section, we present various novel heuristics to swap the assigned requests across vehicles based on the existing assignments. They can be seamlessly integrated into the high-level algorithmic scheme we have discussed so far. While some heuristics demand more computational effort, they are also more powerful in discovering new high-quality assignments. Fundamentally, they are all \textit{local search} heuristics for the epoch-wise \gls{a:rap}. 

\subsection{Multi-Round Linear Assignment (LA-MR)}
We begin by introducing the \gls{a:la-mr} algorithm, a natural extension to \gls{a:la}. Though no swapping is involved, \gls{a:la-mr} forms a foundational framework for other swapping heuristics. Motivated by \gls{a:la} where unserved non-expiring requests can be carried over from one epoch to the next, \gls{a:la-mr} instead performs multiple rounds of bipartite matching within the same epoch. 

The first step is to construct the same bipartite graph $B = (\mathcal{V}, \mathcal{R}_e \cup \hat{\mathcal{R}}, \mathcal{F}_e)$ as that in \gls{a:la}. For simplicity, throughout Section \ref{sec:swap}, we slightly abuse notation by denoting this graph as $B = (\mathcal{V}, \mathcal{R}, \mathcal{F})$. Then, in each round, we perform a constrained bipartite matching such that, if a request $r$ is assigned to a vehicle, any other requests that could have shared a ride with $r$ must not get assigned in the same round. Two requests that can get assigned in the same round are referred to as \textit{independent requests}. This notion is formally shown in Definition \ref{def:independent_requests}. Integrating such a constraint into an optimal bipartite matching problem in each round is non-trivial and overly greedy since we have multiple rounds. Hence, we solve the same \gls{a:ilp} (\ref{eq:la_ilp}) as that in \gls{a:la}, and then sample edges from the optimal matching one by one in an arbitrary order. All edges sampled are considered fixed assignments in this epoch. The sampling procedure must guarantee that all requests are independent. One round finishes once all edges in the current optimal matching are visited. At the end of each round, the algorithm updates the bipartite graph $B$ by removing sampled edges and their corresponding requests (but not vehicles). The algorithm terminates when no further assignments can be made (i.e., no more rounds are needed). Trivially, the algorithm runs in a polynomial time assuming the usual access to the \gls{a:ctsp} oracle.
\vspace{1em}
\begin{definition}[Independent Requests]
    \label{def:independent_requests}
    Assume $B = (\mathcal{V}, \mathcal{R}, \mathcal{F})$ is a bipartite graph capturing a vehicle set, a request set, and a set of feasible candidate trips. Given two requests $r, \; r^{\prime} \in \mathcal{R}$. If there exists a vehicle $v$ such that $(v, r) \in \mathcal{F}$ and $(v, r^{\prime}) \in \mathcal{F}$, we say $r$ and $r^{\prime}$ are dependent on $v$. Otherwise, we call $r$ and $r^{\prime}$ independent requests. 
\end{definition}

A toy example is shown in Figure \ref{fig:la-mr} to illustrate the procedure of \gls{a:la-mr} step by step. The system deploys three vehicles and receives five requests in a decision epoch. In round 1, all feasible trips given by the \gls{a:ctsp} oracle are represented by the edges, and the solid ones form the optimal matching given by \gls{a:ilp}(5). Since requests $2$ and $3$ are dependent on vehicle $1$, they cannot be assigned simultaneously in this round. Suppose the algorithm samples edges $2-1$ and $4-3$, as highlighted in red, and proceeds to the next round. The sets next to the vehicle nodes indicate the current assignment at the end of a round. In round 2, requests $2$ and $4$ are no longer part of the bipartite graph as they are assigned. Similarly, requests $1$ and $3$ are dependent on vehicle $1$, and suppose edge $3-2$ is the one that is not sampled. In round 3, only request $3$ remains in the system and it is ultimately assigned to vehicle $1$ since it incurs a lower cost than that of being assigned to vehicle $2$. The final assignments for vehicles $1$ and $3$ are $\{1, 2, 3\}$ and $\{4, 5\}$ respectively. Vehicle $2$ remains idle.
\input{generator/la-mr}


We emphasize that \gls{a:la-mr} is distinct from executing multiple rounds of \gls{a:la} naively. The former only fixes a subset of the optimal matching (i.e., a partial solution) to the \gls{a:la} solution in each round, whereas the latter accepts the entire optimal matching, and proceeds to the next round if the number of requests is larger than the number of available vehicles. This nuance breaks the artificial constraint between the assignments as discussed in Section \ref{sec:la}. For instance,  consider two travelers with nearly identical origins and destinations who request rides at almost the same time. The most efficient solution would be to place them in the same vehicle, but \gls{a:la}, even when executed naively over multiple rounds, fails to achieve this as either traveler is inevitably placed in the second available vehicle which is possibly very far away. Such an undesirable situation can be avoided in \gls{a:la-mr} as the same vehicle can be assigned to both travelers across two different rounds.

\subsection{Multi-Round Linear Assignment with Naive Swapping (LA-MR-NS)} \label{sec:la-mr-ns}

In this section, we introduce our first swapping heuristic that adds an additional layer on top of the \gls{a:la-mr} algorithm in a naive way that preserves the efficiency of the bipartite matching. We call the modified algorithm \gls{a:la-mr-ns}. To better illustrate the heuristic, we define a new notion called \textit{naive swap} shown in Definition \ref{def:delta-swap}.
\begin{definition}[Naive Swap] \label{def:delta-swap}
    Given two trips $f_1 = (v_1, \mathcal{R}_1 \cup \{r\})$ and $f_2 = (v_2, \mathcal{R}_2)$, with $v_1 \neq v_2$ and $r \notin \mathcal{R}_2$, a naive swap $\delta(r, \, f_1 \rightarrow f_2)$ reassigns request $r$ from trip $f_1$ to trip $f_2$, resulting in two new trips $f_1^\prime = (v_1, \mathcal{R}_1)$ and $f_2^\prime = (v_2, \mathcal{R}_2 \cup \{r\})$. The associated cost reduction is $c(r, \, f_1 \rightarrow f_2) = c(f_1^\prime) + c(f_2^\prime) - c(f_1) - c(f_2)$. A naive swap with a positive cost reduction is valid.
\end{definition}
We start with the same bipartite graph as presented in \gls{a:la-mr}. The first round is identical to that in \gls{a:la-mr}. From the second round onward, for every request that has already been assigned in the previous rounds, a new duplicate node is added to the graph representing the request. Furthermore, a new edge is added between this new node and any vehicle if (1) the request can be feasibly reassigned to the vehicle, and (2) the reassignment results in a positive cost reduction. Each such edge represents a valid naive swap of reassigning the request from its currently assigned vehicle to the new vehicle. Isolated duplicate nodes connecting to no vehicle can be discarded afterward. By this construction, we obtain an extended bipartite graph $B^\prime = (\mathcal{V}, \mathcal{R} \cup \mathcal{R}_{\delta}, \mathcal{F} \cup \mathcal{F}_\delta)$, where $\mathcal{R}_{\delta}$ and $\mathcal{F}_{\delta}$ are the set of newly added nodes and edges, respectively. Then, in every round of \gls{a:la-mr-ns}, we solve the same \gls{a:ilp} (\ref{eq:la_ilp}) as that in \gls{a:la} and \gls{a:la-mr}, but on the extended bipartite graph $B^\prime$ instead of the original one $B$. The weight of every edge in $\mathcal{F}_{\delta}$ is the cost reduction of the corresponding naive swap, which can be computed by the \gls{a:ctsp} oracle as well. 

Similar to that in \gls{a:la-mr}, we sample edges (i.e., partial solution) one by one in an arbitrary order from the optimal matching once we solve \gls{a:ilp} (\ref{eq:la_ilp}) in a round. In the \gls{a:la-mr-ns}, we are no longer restricted by the notion of independent requests. Instead, we introduce another notion called \textit{independent vehicles} shown in Definition \ref{def:independent_vehicles} to facilitate the sampling procedure. Two vehicles are considered independent if there is no request currently assigned to one of the vehicles that can feasibly be assigned to the other. The sampling must guarantee that all naive swaps only reassign requests from or to independent vehicles. In other words, no pair of sampled edges is allowed to connect to dependent vehicles. Again, at the end of each round, the algorithm removes the sampled edges and their corresponding request nodes and proceeds to the next round. The algorithm terminates when no new assignment can be made.
\begin{definition}[Independent Vehicles]
    \label{def:independent_vehicles}
    Assume $B^\prime = (\mathcal{V}, \mathcal{R} \cup \mathcal{R}_{\delta}, \mathcal{F} \cup \mathcal{F}_{\delta})$ is an extended bipartite graph. Given two distict vehicles $v, \; v^\prime \in \mathcal{V}$, if there exists a naive swap $\delta = (r, \, f \rightarrow f^\prime) \in \mathcal{F}_{\delta}$ such that $r \in \mathcal{R}_{\delta}$, $v \in f$ and $v^\prime \in f^\prime$, we call $v$ and $v^\prime$ are dependent. Otherwise, we call $v$ and $v^\prime$ independent vehicles.
\end{definition}

A toy example is shown in Figure \ref{fig:la-mr-ns} to illustrate the procedure of \gls{a:la-mr-ns} step by step. The system deploys three vehicles and receives four requests in a decision epoch. In round 1, all feasible trips given by the \gls{a:ctsp} oracle are represented by the edges, with the solid ones forming the optimal matching given by \gls{a:ilp}(5). At this stage, all vehicles are independent since they do not have any passengers onboard. Thus, all edges are sampled for assignment, as highlighted in red, and the algorithm proceeds to the next round. The sets next to the vehicle nodes indicate the current assignment at the end of a round. In round 2, for each assigned request except request $1$, a duplicate node, shown as a diamond, is added back to the bipartite graph. Each edge connecting such a node represents a valid (i.e., negative-cost) naive swap that reassigns the corresponding request to a different vehicle. In this round, edges $2-2$ cannot sampled together with either edge $1-1$ or edge $3-3$ because vehicle pair $(1, 2)$ and depends on request $2$ and vehicle pair $(2, 3)$ depends on request $3$. Suppose the algorithm samples edges $1-1$ and $3-3$ and continues. In round 3, suppose only edges $2-2$ and $3-1$ are left. Note that a naive swap is defined regarding two trips, not two vehicles, meaning that an edge appearing in a round might not exist in the next round, or vice versa, due to an update on the vehicle route. The algorithms ultimately sample edge $2-2$ instead of $3-1$, leading to the final assignment. The termination step is not shown in the diagram: the algorithm must check that, after round 3, there is no edge generated.  
\input{generator/la-mr-ns}

The justification for the independent vehicles is two-fold: (1) Multiple naive swaps to dependent vehicles in the same round may result in a worse solution to the epoch-wise \gls{a:rap} locally, as naive swaps do not automatically guarantee a strict decrease in the cost of the request-vehicle assignment even though when the cumulative cost reduction of those naive swaps is positive. Figure \ref{fig:la-mr-ns-2} illustrates such an example. (2) Unlike the case of \gls{a:la-mr} where request independence is used as a heuristic for good assignments, we need vehicle independence to ensure the algorithm can terminate in the case of \gls{a:la-mr-ns}. Vehicle dependence guarantees a decreasing cost over rounds, avoiding the algorithm looping back to a previously visited solution. In short, such a characteristic is the core reason why we call the swapping naive --- each swap calls the \gls{a:ctsp} oracle separately in a decoupled manner, and thus the accurate cost reduction of assigning a request to a vehicle cannot be obtained if another request is simultaneously being removed from it via swapping. It further motivates us to study a non-naive swapping heuristic, which will be discussed in \ref{sec:la-mr-ps}.
\input{generator/la-mr-ns-2}

\subsection{Multi-Round Linear Assignment with Proper Swapping (LA-MR-PS)} \label{sec:la-mr-ps} 
In this section, we propose another \gls{a:la-mr}-based swapping heuristic whose cost reduction respects the accurate objective of the epoch-wise \gls{a:rap}. We call the modified algorithm \gls{a:la-mr-ps}. It avoids the issue of cost reduction being computed in a decoupled way by converting the model from bipartite matching to general matching. This change preserves the polynomial time complexity as general matching is not an NP-hard problem.

Again, we start with the bipartite graph used in \gls{a:la-mr} and \gls{a:la-mr-ns} in each round. Since every swap of a request involves two vehicles, we extend the graph by adding an edge connecting every pair of vehicle nodes if there is at least one request in one of the vehicles that can feasibly be reassigned to the other and results in a cost reduction. This breaks the structure of bipartite matching because we have a new set of edges, denoted by $\mathcal{F}_{\gamma}$, involving vehicle nodes only. Each such edge corresponds to a \textit{proper swap} which is formalized in Definition \ref{def:proper-swap}. Since every proper swap represents an optimal naive swap between any vehicle pairs, these edges can be viewed as directed pointing from the source vehicle to the target vehicle. Once the non-bipartite graph $B^\prime = (\mathcal{V} \cup \mathcal{R}, \mathcal{F} \cup \mathcal{F}_{\gamma})$ is constructed, we solve the General Maximum Weight Matching Problem on $B^\prime$. In theory, many commonly used algorithms in bipartite matching, such as Edmond's algorithm and the primal-dual algorithm, can be generalized to solve this problem. In our implementation, we adopt its \gls{a:ilp} formulation and feed it to a \gls{a:mip} solver as it is sufficiently efficient. Next, we sample edges from the optimal matching one by one in an arbitrary order and respect the request independence in Definition \ref{def:independent_requests}. This sampling procedure is identical to that in \gls{a:la-mr}. Thus, it is part of the heuristic rather than a termination condition as vehicle independence in \gls{a:la-mr-ns}. Similar to other \gls{a:la-mr}-based algorithms, \gls{a:la-mr-ps} proceeds to the next round after sampling and terminates if no new assignment can be made.
\begin{definition}[Proper Swap] \label{def:proper-swap}
Given two vehicles $v_1$ and $v_2$ and their corresponding trips $f_1 = (v_1, \mathcal{R}_1)$ and $f_2 = (v_2, \mathcal{R}_2)$, identify two subsets of requests $\mathcal{R}^{\gamma}_1 \subseteq \mathcal{R}_1$ and $\mathcal{R}^{\gamma}_2 \subseteq \mathcal{R}_2$ such that (1) for every $r \in \mathcal{R}^{\gamma}_1$, there is a naive swap $\delta(r, \, f_1 \rightarrow f_2)$, and (2) for every $r \in \mathcal{R}^{\gamma}_2$, there is a naive swap $\delta(r, \, f_2 \rightarrow f_1)$. A proper swap $\gamma(v_2, v_2)$ is the most cost-efficient valid naive swap $\delta^*$ given by 
\[
\delta^* = \arg\min\limits_{\delta}\{ \min\limits_{r \in \mathcal{R}^{\gamma}_1} c(r, \, f_1 \rightarrow f_2), \; \min\limits_{r \in \mathcal{R}^{\gamma}_2} c(r, \, f_2 \rightarrow f_1)\}
\]
If $\mathcal{R}^{\gamma}_1 = \mathcal{R}^{\gamma}_2 = \varnothing$ there is no proper swap between $v_1$ and $v_2$.
\end{definition}

A toy example is shown in Figure \ref{fig:la-mr-ps} to illustrate the procedure of \gls{a:la-mr-ps} step by step. The setup is identical to the example in Figure \ref{fig:la-mr-ns} for \gls{a:la-mr-ns}. In round 1, three request-vehicle assignments are suggested by linear assignment but only two of them (e.g., edges $2-1$ and $4-3$) are sampled to maintain request independence. In round 2, an extra directed edge pointing from vehicle $1$ to vehicle $2$ is added to represent a proper swap between them. Cost-increasing proper swaps (i.e., invalid proper swaps) are not shown in the figure. Suppose the algorithm samples edge $1-1$ based on the optimal matching (e.g., edges $1-1$ and $3-3$). In round 3, the only remaining request (i.e., request $3$) is assigned to vehicle $3$ while request $2$ is swapped from vehicle $1$ to vehicle $2$. Though the final assignment is identical to that in \gls{a:la-mr-ns}, the solution process is noticeably different in this case.
\input{generator/la-mr-ps}

Obviously, \gls{a:la-mr-ps} needs multiple \gls{a:ctsp} queries to determine every proper swap, and thus it is more computationally demanding than \gls{a:la-mr} and \gls{a:la-mr-ns} as they only need one \gls{a:ctsp} query for each edge. On the other hand, the algorithm is more powerful as it explores a large solution space via swapping while ensuring a strict improvement in objective value at each round.

\subsection{Multi-Round Linear Assignment with Cyclic Exchange (LA-MR-CE)}

In this section, we introduce another method, \gls{a:la-mr-ce}, that pushes further into the trade-off between computation time and solution quality by adding a more sophisticated swapping mechanism called cyclic exchange. Originating from one of the Very Large Neighborhood Search (VLNS) heuristics proposed by \cite{ahuja2000very}, it has the ability to swap multiple request-vehicle assignments simultaneously. Unlike \gls{a:la-mr-ns} and \gls{a:la-mr-ps}, the \gls{a:la-mr-ce} is a two-stage algorithm that begins by assigning new requests using \gls{a:la-mr} and performs cyclic exchange at the second stage. Still running in polynomial time, the algorithm incurs extra computation burden in cyclic exchange compared to single swapping. However, it may be an attractive feature to encourage this under the hypothesis that swapping earns most of the marginal benefit within many ride-pool assignment heuristics.

The cyclic exchange scheme is designed to address the general partition problem, where we have a universe of elements and a set of feasible partitions, each associated with a cost. The objective is to find a minimum-cost allocation of each element to a unique partition. The pattern of cyclic exchange is visualized in Figure \ref{fig:ce}, where C1 and C2 are two possible cycles. In its original definition, exchanges must form a cycle, and chains are not allowed. In the context of the epoch-wise \gls{a:rap}, a vehicle with all requests currently assigned to it is considered a partition, and an empty vehicle is also regarded as a special partition. Given the initial partitioning outputted by \gls{a:la-mr}, our task is to consider all possible one-request swaps across all partitions, forming either a chain or a cycle, to reduce the total cost of the solution.  It is important to note that some requests might remain unassigned when \gls{a:la-mr} is completed. In such cases, these requests are assigned to a special partition representing the ``null'' vehicle. The cost reduction for moving any request from the ``null'' partition to another partition is set to a large positive value $U$, while the reverse operation is assigned a cost reduction of $-U$. This ensures that maximizing \gls{a:sr} takes precedence over minimizing \gls{a:vmt}.
\input{generator/ce}

To perform cyclic exchange on our problem, we first construct a graph $G_{CE} = (N_{CE}, A_{CE})$ with a node for each request and a dummy node for each vehicle. For any two distinct requests $r_i$ and $r_j$, we add a directed edge connecting $r_i$ to $r_j$ if they are assigned to different vehicles and it is feasible to route the vehicle currently assigned to $r_j$ if $r_j$ were replaced by $r_i$. Additionally, we add edges between requests and vehicles representing simple assignment addition or removal without replacement. A directed edge, representing assignment addition, connects request $r$ to vehicle $v$ if it is feasible to route $v$ with $r$ added as an assignment. Conversely, a directed edge connects vehicle $v$ to request $r$ if $r$ is unassigned or assigned to a vehicle other than $v$. It represents removing $r$ from its current vehicle assignment or changing it from unassigned to assigned, as would happen for the first request in a chain of exchanges. For all edges, the associated weight is the cost reduction experienced by the vehicle. Including dummy vehicle nodes and their connecting edges is necessary to transform a chain in our problem into a cycle in graph $G_{CE}$. Any edge pointing from a dummy vehicle node and a request serves as the final piece to link the head and the tail of a chain. With this modified graph representation, the algorithm developed for the original cyclic exchange can be utilized directly. An example of graph $G_{CE}$ is visualized in Figure \ref{fig:la-mr-ce}.
\input{generator/la-mr-ce}

The crux of cyclic exchange is identifying valid cycles in the graph $G_{CE}$, which are defined as follows. The first validity restriction adheres to the original definition of cyclic exchange which avoids swapping more than one request out of the same partition. This constraint helps maintain computational tractability by limiting the search space. More importantly, this ensures the validity of the objective function, as the value gained by moving multiple requests out of the same partition is not equal to the sum of the values gained by removing them individually. The second restriction on vehicle nodes is for cycle simplification, as any cycle containing more than one vehicle node can be decomposed into multiple smaller cycles, each with only one vehicle node, while maintaining the same total cost reduction. For example, in Figure \ref{fig:la-mr-ce}, selecting the cycle $v_1-r_2-v_3-r_1-r_3-v_1$ is equivalent to selecting cycles $r_2-v_3-r_2$ and $v_1-r_1-r_3-v_1$ sequentially.
\begin{definition}[Valid Cycle] \label{def:valid_cycle}
    A cycle in $G$ is considered valid if (1) no two request nodes in the cycle are associated with the same vehicle, and (2) at most one node in the cycle is a vehicle node. The associated cost of a cycle is the total cost reduction of swapping all involved requests simultaneously. 
\end{definition}

Next, we introduce an algorithm to perform cyclic exchange.  It requires the use of two additional data structures: a frontier set $P$ and an associative array $M$.  Each key-value pair in the associative array $M$ maps a node to a set of nodes, which is used to add some nodes back to $P$ during the algorithm's execution. Initially, $P$ contains all request nodes that are not dummy vehicle nodes, and $M$ maps all keys to empty sets. In each iteration, an arbitrary node $i$ is removed from $P$ to search for a maximum cost-reducing valid cycle starting at $i$.  If such a cycle is found, we execute all request swaps denoted by the edges in the cycle, possibly creating or destroying some edges as appropriate, and update the weights on all affected remaining edges.  Also, we add $i$ back to $P$ to explore again in case there are more cost-reducing cycles to be found. However, if no such cycle originating from $i$ is found, we instead use $M$ to store the set of nodes that were explored by the cycle-search algorithm, denoted by $\mathcal{S}$. As long as no new arcs are connected to any node in $\mathcal{S}$ and none of the existing arcs adjacent to nodes in $\mathcal{S}$ have their weights increased, then it is safe to assume that any search from $i$ would continue to return no improving cycle. If, in any future iteration, the graph is updated with the set of affected nodes denoted by $N_A$, we add back to the frontier $P$ any request node $j$ such that $M[j] \cap N_A \neq \varnothing$, as searches for cycles from these nodes may now yield valid improvements, even if they did not in earlier iterations. The algorithm terminates when $P$ is empty.

The sub-routine of finding a maximum cost-reducing cycle is equivalent to finding the maximum weight cycle in a simple graph, which is known to be NP-hard. Unlike the minimum weight version that can be transformed into the shortest path problem, the maximum weight path/cycle problem has no optimal substructure for any graph. Thus, any Dijkstra-like polynomial-time algorithm will fail. To obtain high-quality cycles efficiently, we utilize a node-labeling search algorithm that, at each frontier node being processed, records the objective of all possible valid paths from the source node to this node. A valid path shares the same definition as a valid cycle, as shown in Definition \ref{def:valid_cycle}, except that it is a path. Furthermore, we adopt the following two heuristics to make the algorithm computationally tractable. In short, the complete version of \gls{a:la-mr-ce} is shown in Algorithm \ref{alg:ce}.
\begin{enumerate}
    \item Without loss of generality, we require that the cumulative sum of cost reduction along any exploring path be strictly greater than the negative of the potential cost reduction from removing the source request node from its vehicle assignment. This heuristic is justified since the cumulative cost reduction may be negative (i.e., the routing cost could increase), but we aim to avoid intermediate swaps that lead to excessive cost increases, even though small increments are permissible during the process.
    \item As suggested by \cite{ahuja2000very}, each time we process a node, if a path has the best objective over all paths that have visited this node previously, we discard all old paths and save the newest one only. This heuristic prevents the algorithm from tracking an exponential number of paths in the search. To improve the solution quality, one can specify a hyper-parameter to reserve more than one path.  
\end{enumerate}

\begin{algorithm}[!htbp]
\caption{Multi-Round Linear Assignment with Cyclic Exchange}
\label{alg:ce}
\begin{algorithmic}[1]
\setlength{\baselineskip}{1.1\baselineskip} 
\State Construct $G_{CE} = (N_{CE}, A_{CE})$ from the assignment $\bm{x}$ given by \gls{a:la-mr}
\State $M \gets \{ (M[i]: \varnothing) \mid i \in N_{CE}\}$
\State $N_R \gets \{i \in N_{CE} \mid i \notin \mathcal{V}\}$
\State $P \gets N_R$

\While{$P \neq \varnothing$}
    \State $i \gets P.\textsc{Pop}()$ \Comment{Any arbitrary order}
    \State $(\mathcal{C}, \mathcal{S}) \gets \textsc{MaxCostReductionCycle}(G_{CE}, i)$
    \If{$\mathcal{C} = \varnothing$}
        \State $M[i] \gets \mathcal{S}$  
        \State \textbf{continue}
    \EndIf
    \State $P \gets P \cup \{i\}$
    \State $\bm{x} \gets \textsc{UpdateAssignment}(\bm{x}, \mathcal{C})$
    \State $G'_{CE} \gets \textsc{UpdateGraph}(G_{CE}, \mathcal{C})$
    \State $N_a \gets \textsc{AffectedNodes}(G_{CE}, G'_{CE})$ \Comment{Nodes affected by graph update}
    \State $G_{CE} \gets G'_{CE}$
    \For {$j \in N_R$}
        \If {$M[j] \cap N_a \neq \varnothing$}
            \State $P \gets P \cup \{j\}$
        \EndIf
    \EndFor
\EndWhile
\State \textbf{return} $\bm{x}$

\vspace{0.5em}

\Procedure{MaxCostReducingCycle}{$G_{CE}$, $n$} 
\State $T \gets \text{cost reduction by removing request at } n \text{ from its vehicle without replacement}$
\State $\mathcal{S} \gets \{n\}$ \Comment{Nodes explored}
\State $\mathcal{V} \gets \{(i: -\infty) \mid i \in N_{CE}\}$ \Comment{Cumulative sum of cost reduction}
\State $\mathcal{V}[n] \gets 0$
\State $\mathcal{P} \gets \{ n : \varnothing\}$ \Comment{The best partial paths by endpoint}
\State $\mathcal{Q} \gets \{n\}$ \Comment{A priority queue of the frontier to explore}
\While {$\mathcal{Q} \neq \varnothing$} 
    \State $i \gets \textsc{QueuePopMax}(\mathcal{Q}, \mathcal{V})$
    \If {$\mathcal{V}[i] < -T$}
        \State \textbf{break}
    \EndIf
    \For {$j \in \textsc{Neighbors}(i)$}
        \If {$\textsc{NotFeasible}(\mathcal{P}[i] \cup (i, j))$}
            \State \textbf{continue}
        \EndIf
        \State $\mathcal{S} \gets \mathcal{S} \cup \{j\}$
        \State $v \gets \textsc{Value}(\mathcal{P}[i] \cup (i, j))$
        \If {$v > \mathcal{V}[j]$}
            \State $\mathcal{V}[j] \gets v$
            \State $\mathcal{P}[j] \gets \mathcal{P}[i] \cup (i, j)$
            \If {$j \neq n$}
                \State $\mathcal{Q} \gets \mathcal{Q} \cup \{j\}$
            \EndIf
        \EndIf
    \EndFor
\EndWhile
\If{$\mathcal{V}[n] > 0$}
    \State \textbf{return} $(\mathcal{P}[n], \mathcal{S})$
\Else
    \State \textbf{return} $(\varnothing, \mathcal{S})$
\EndIf
\EndProcedure

\end{algorithmic}
\end{algorithm}

\section{\gls{a:ctsp} Algorithms and Solution Stability} \label{sec:stability}
Regardless of the algorithm used to make the request-vehicle assignment, the \gls{a:ctsp} oracle must frequently re-compute feasible vehicle routes and the associated costs. A \gls{a:ctsp} oracle $\mathcal{H}$ is a function that takes a vehicle route and a set of requests as the input and returns an updated route. Re-computation is necessary at every decision epoch so that the \gls{a:rap} can consider ways to update each vehicle's assignments given the arrival of new requests, with different algorithms requiring a different number of calls to oracle at each epoch.

The \gls{a:ctsp} is NP-hard as it generalizes the \gls{a:tsp}. When the number of requests assigned to a vehicle is relatively small, an \textit{exact recursive search} algorithm is usually sufficient to find the optimal routes. In such a case, we first extract all nodes associated with the current vehicle route and new requests. Starting with each node, we build a search tree in a depth-first search manner, by adding one node to a sub-route at a time if feasible. After recursively enumerating all possible routes, the algorithm returns the least-cost route. To improve search efficiency, we also maintain a list of \textit{follower nodes} for each node in our implementation. They represent the nodes that can only be visited after the target node. Examples include that a drop-off node must be visited after the corresponding pick-up node from the same request, and a node that cannot be reached before 9:00 p.m. must be visited after a node that must be visited by 8:55 p.m. in any feasible solution. Whenever we extend a sub-route, we only consider adding nodes that do not violate the precedence constraints. Exploiting such precedence constraints between nodes avoids redundant feasibility checks and thus speeds up the algorithm.

However, as the number of requests grows, it becomes computationally intractable or excessively costly to explore all sub-routes. Therefore, heuristics are often employed for larger problem instances. Despite their convenience, a heuristic that successfully returns a route may fail to regenerate a consistent (or even a feasible) sub-route starting from later in the trip even in the absence of new requests. At its most naive, a heuristic might not replicate the solution if the same \gls{a:ctsp} instance is re-computed. In this sense, the heuristic is unstable. The issue of \textit{solution stability} is crucial in the implementation of all ride-pool assignment algorithms to yield high performance. We are concerned that an unstable heuristic, compared to a stable one, is less likely to find a solution when new requests are added in future epochs. Moreover, unstable heuristics may generate suboptimal vehicle routes with highly varying costs upon re-computation, undermining the effectiveness of swapping-based algorithms. These algorithms heavily rely on the precise and stable computation of swapping costs, and drastic changes in swapping costs across different rounds can impact their termination. We
formally define the stability of a heuristic in Definition \ref{def:stable_heuristic}.

Additionally, any heuristic can be used in conjunction with a \textit{threshold policy} to create a new heuristic such that the target heuristic is only invoked when a pre-defined quantity, as a function of the \gls{a:ctsp} input, exceeds a threshold. Otherwise, the exact recursive search algorithm is used. As a practical example, a threshold can be determined by the total number of requests involved. This simple scheme provides more flexibility in the \gls{a:ctsp} computation as the size of instances dynamically changes across rounds and epochs, allowing us to take advantage of exact computation or the speed
of a heuristic as as appropriate. Such a threshold-enabled heuristic is stable only if it is built on a stable heuristic and the pre-defined quantity is a monotonic increasing function of the number of vehicle stops involved in the \gls{a:ctsp} input. In the following subsections, we will introduce some practical heuristics and discuss their solution stability.
\begin{definition}[Stable Heuristic] \label{def:stable_heuristic}
Given any feasible vehicle route $S_v$, any set of new coming requests $\mathcal{R}$, and a transition $\mathcal{T}$ on $S_v$ after any time elapsed $\Delta t \geq 0$, a \gls{a:ctsp} heuristic $\mathcal{H}$ is stable if the original route $S_v$ and the updated route $\mathcal{H}(\mathcal{T}(S_v, \Delta t), \mathcal{R})$ have the same ordering for their common nodes. 
\end{definition}

\subsection{Insertion Heuristic}
The insertion heuristic is the most naive heuristic commonly used in the \gls{a:tsp} and was the choice used in the original work of the \gls{a:rtv} algorithm \cite{alonso2017demand}. It is a construction heuristic that starts from an empty set or a partial solution and inserts additional elements either one by one or group by group in an arbitrary order sequentially. At each step, the algorithm preserves the order of all previously added elements. In our implementation, this heuristic starts with an empty vehicle route and inserts one arbitrary node at a time at the optimal feasible position in the route until (1) all nodes are added, or (2) no feasible insertion can be found.

\begin{figure}[htbp!]
    \centering
    \begin{subfigure}[b]{0.45\textwidth}
        \includegraphics[trim={210 442 210 113},clip]{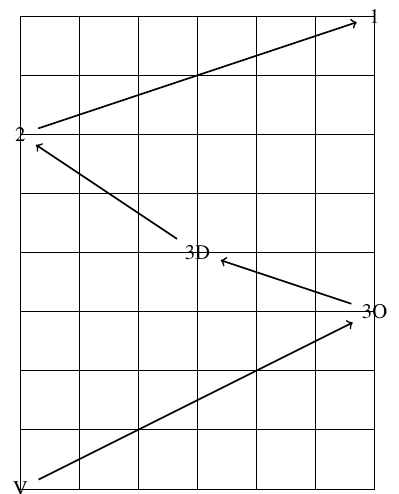}
        \caption{The initial feasible vehicle route}
        \label{fig:unstable_ctsp_initial}
    \end{subfigure}
    \begin{subfigure}[b]{0.45\textwidth}
        \includegraphics[trim={210 109 210 446},clip]{figures/ctsp.pdf}
        \caption{The infeasible shortest vehicle route in re-computation}
    \end{subfigure}
    \caption{An instance demonstrating the solution instability of insertion heuristic}
    \label{fig:unstable_ctsp}
\end{figure}

Unfortunately, the naive insertion heuristic is unstable as shown in the example depicted in Figure \ref{fig:unstable_ctsp}. Suppose there is an existing vehicle route with two scheduled drop-off nodes, indexed $1D$ and $2D$, and a new incoming request with its origin and destination indexed $3O$ and $3D$ respectively. Suppose all drop-off nodes must be visited by time $20$ and $3O$ must be visited by time $10$ due to the time constraint. This \gls{a:ctsp} instance is visualized in Figure \ref{fig:unstable_ctsp_initial}, where the grid map is in a 2-D Euclidean space and each cell is $1 \times 1$. The vehicle speed is assumed to be $1$ as well and all quantities are unitless. One feasible solution given by the insertion heuristic is Route $\{V, 3O, 3D, 2, 1\}$. Its associated cost is $19.8$ units, which is indeed optimal among all possibilities. After the vehicle reaches $3O$, suppose we need to re-compute the vehicle route using the insertion heuristic. Since the order of the recursive search is arbitrary, the heuristic may start by selecting node $1D$ and $2D$ sequentially, creating a sub-route with the order of $1D$ and $2D$ flipped compared to the previous solution. Next, node $3D$ needs to be inserted into the sub-route, but none of the insertion positions yield a feasible route since the remaining total travel time, plus the time already traveled, exceeds 20 units. The shortest path $\{V, 1, 2, 3D\}$ gives a total travel time of $21.6$ as shown in Figure \ref{fig:unstable_ctsp}.

In short, the inversion of nodes $1D$ and $2D$ violates Definition \ref{def:stable_heuristic}, indicating that the insertion heuristic is unstable. More importantly, this instability, which prevents the consistent regeneration of a part of the solution, is the root cause leading to the infeasible insertion of node $3D$ in the subsequent time step. This argument shows that heuristic stability is a more nuanced concept than the simple fact that any heuristics cannot always find feasible solutions to the \gls{a:ctsp} given its NP-hardness. We emphasize that such instability may further diminish a heuristic's ability to generate feasible vehicle routes with relatively consistent costs compared to other stable heuristics.

\subsection{On-board Order Fixing (OOF) Heuristic} \label{sec:oof}
The \gls{a:oof} heuristic is introduced in \cite{alonso2017demand} as a threshold-enabled heuristic. When the number of passengers onboard plus the number of requests yet to board is larger than a threshold, it fixes the order of drop-off nodes of all onboard passengers and inserts both pick-up and drop-off nodes of yet-to-board requests in any order leading to the best solution. In other words, the solution is first initialized to be the sub-route with only the drop-off nodes ordered as it was in the current vehicle route, making it distinct from the naive insertion heuristic. According to Definition \ref{def:stable_heuristic}, \gls{a:oof} is stable.

\subsection{Limit and Recall Prefix (LRP) Heuristic} \label{sec:lrp}
The \gls{a:lrp} heuristic is a generalization of the \gls{a:oof} heuristic designed to handle larger \gls{a:ctsp} instances. The fundamental idea is that if a vehicle has been assigned many requests, adding new requests in a future iteration is more likely to alter the later portions of the vehicle route rather than the imminent ones. Similarly, \gls{a:lrp} is a threshold-enabled heuristic. When the total number of nodes, due to passengers onboard and requests yet to board, exceeds a threshold $\eta \in \mathbb{N}_+$, the heuristic is triggered. Note that each onboard passenger only contributes one node (i.e., the drop-off node) while each yet-to-board request contributes two nodes (i.e., a pick-up node and a drop-off node). Upon invocation, the heuristic recalls the vehicle route computed in the previous iteration from memory and removes all nodes (if applicable) already visited by the beginning of the current iteration. An iteration can be a round in multi-round algorithms, such as swapping algorithms discussed in Section \ref{sec:swap}, or a decision epoch in single-round algorithms, such as the \gls{a:rtv} algorithm discussed in Section \ref{sec:algos}. Next, a prefix $\overline{S} \subseteq S$ is identified in the remaining sub-route $S$, such that the number of nodes in the suffix $\underline{S} = S \setminus \overline{S}$, plus the number of nodes from new requests $\mathcal{R}$ in this iteration, equals the threshold (i.e., $|\underline{S}| + |\mathcal{R}| = \eta$). If the number of nodes from new requests exceeds the threshold (i.e., $|R| > \eta$), the heuristic returns no solution. Otherwise, an exact recursive search is employed to optimally insert all nodes from $\underline{S} + \mathcal{R}$ into the prefix $\overline{S}$ while keeping the order of $\overline{S}$ intact. The \gls{a:lrp} heuristic is also stable according to Definition \ref{def:stable_heuristic}.

\section{Open-Source Implementation} \label{sec:codebase}

Our implementation of the aforementioned ride-pool assignment algorithms is open-sourced at \hyperref[https://github.com/MAS-Research/RidepoolSimulatorAssignmentAlgs]{https://github.com/MAS-Research/RidepoolSimulatorAssignmentAlgs}. The algorithms form the core module of a complete pipeline capable of simulating a ride-pooling system with user-specified input and control parameters. The codebase, written in C++, is highly optimized to support very large-scale systems in high-density urban areas. The underlying \gls{a:mip} solver utilizes SCIP APIs \cite{BolusaniEtal2024OO} while the \gls{a:lp} solver utilizes COIN-OR Clp.

The choice of C++ as the programming language was made for its efficiency in handling large computational demands, especially for algorithms requiring the construction of large \gls{a:rtv} graphs. C++ enables modular program development, with clear variable scoping rules that facilitate program verification through decomposition. It also provides a high degree of control over data structures, preventing unnecessary duplication of data when passing arguments to functions. The typing system in C++ ensures clarity throughout the program regarding the purpose and functionality of various functions.

The codebase is organized into four major directories: \texttt{data} contains user-provided input data for the ride-pooling system, such as the map, requests, and vehicles; \texttt{src} contains all source files in the \texttt{.cpp} format; \texttt{headers} contains all the header files corresponding to each source file; \texttt{threadpool} contains necessary files to support multi-threading in C++. The output of the program, which will be automatically generated in a user-specified directory, includes log files from each decision epoch and key statistics of the full-day simulation such as the service rate, shared rate, and total \gls{a:vmt}. Detailed explanations of some important components in \texttt{src} are as follows. For a complete description and usage of each component, refer to \texttt{README.md} in the repository.
\begin{itemize}
    \item \texttt{algorithms:} This is a level-2 directory that includes all source files for the ride-pool assignment algorithms and swapping heuristics discussed in this paper. The file called \texttt{ilp\_common.cpp} is the \gls{a:ilp} solver being called by several relevant algorithms. All other files are named descriptively.
    \item \texttt{settings.cpp:} It contains all macros and tunable hyper-parameters such as request batching interval, maximum detour, computational time limit for the algorithms, and so on. Most of them are linked to input arguments in the \texttt{main.cpp} file and can be specified by users.
    \item \texttt{routeplanner.cpp:} It includes the \gls{a:ctsp} solver and helper functions. When the number of stops in the vehicle route exceeds a limit called \texttt{CTSP\_ENUMERATE\_LIMIT}, the \gls{a:ctsp} heuristics discussed in Section \ref{sec:stability} are used. Otherwise, the solver recursively searches for the optimal route.
    \item \texttt{network.cpp:} It includes functions to initialize a network based on map data for simulation. It also includes an optimized implementation of Dijkstra's algorithm to compute the shortest path between nodes. 
    \item \texttt{rebalance.cpp:} It includes necessary files to perform vehicle rebalancing, with the algorithm discussed in Section \ref{sec:problem}. 
\end{itemize}

\section{Numerical Experiments} \label{sec:experiments}
In this section, we will benchmark the performance of the aforementioned ride-pool matching algorithms in the full-day simulation setting (Section \ref{sec:full_day}) and analyze the correlation of epoch-wise solutions over time (Section \ref{sec:correlation}).

All the experiments are run on an m7i.8xlarge AWS instance with 32 CPU cores (Intel(R) Xeon(R) Platinum 8488C) and 128 GiB memory. The control parameters are set as follows if not specified individually. The number of vehicles is set to $1000$. The capacity of each vehicle is set to $4$. The maximum waiting and detour times are set to 5 minutes and 10 minutes respectively. The request batching interval is set to $60$ seconds. The maximum number of threads is set to $10$ for all components that support parallel computing (e.g., oracle calls for the \gls{a:ctsp} solver). The computational time limit for the fast \gls{a:rtv} algorithm is $10$ seconds. No computational time limit is set for the \gls{a:cg} algorithm. The maximum number of stops for the \gls{a:ctsp} solver to execute an exact recursive search is set to $12$. For every trip with more stops than $12$, the solver uses the \gls{a:oof} heuristic introduced in Section \ref{sec:oof}.

\subsection{Full-Day Simulation} \label{sec:full_day}
We compare the algorithms by running the full-day simulation on an identical dataset. We use one week of request data from 2013-03-01 to 2013-03-07 in the Yellow Taxi Trip Records provided by the NYC Taxi and Limousine Commission (TLC). The underlying map is publicly available in NYC Open Data. We geo-fenced the map to be within Manhattan only. Requests with either an origin or destination outside of the boundary are filtered out. For each day, we uniformly sample data at three levels: full, two-thirds, and one-third to vary the demand stress in the experiments.

Table \ref{tab:service_rate} shows the average full-day service rates for all algorithms at all levels. A box plot of service rate at the full demand level is depicted in Figure \ref{fig:box-plot}. The first thing to note from Table \ref{tab:service_rate} is that the full \gls{a:rtv} algorithm fails to give any result at the full level of demand within $100$ hours due to the excessive computation burden, though it outperforms most other algorithms at the two-thirds and half levels. This is not surprising given that it is an exact method designed to find the optimal solution to the epoch-wise \gls{a:rap}. 

Among other non-exact algorithms based on heuristics, we observe that \gls{a:la} and \gls{a:la-mr} perform almost identically, but consistently get lower service rates than their advanced variants that allow swapping. This suggests that swapping heuristics can boost ride-pool assignment algorithms by encouraging local search on top of the very efficient \gls{a:la} framework. Among the three algorithms with swapping, \gls{a:la-mr-ce} results in the highest service rate. More importantly, \gls{a:la-mr-ce} can even beat the full \gls{a:rtv} algorithm when the demand level is relatively high (with two-thirds demand). \gls{a:la-mr-ns} performs better than \gls{a:la-mr-ps} at the full and two-thirds levels of demand and the performance gap is positively correlated with the demand level, suggesting that \gls{a:la-mr-ns} is more suitable for large-scale systems while \gls{a:la-mr-ps} is more suitable for small to medium systems. 

The fast \gls{a:rtv} algorithm ranks second with one-third demand but becomes less effective than the other algorithms at higher demand levels. On the contrary, \gls{a:cg} performs relatively well when the demand level is very high, but its performance notably drops at lower demand levels. This is likely because it generates fewer high-quality candidate trips at lower demand levels.
\input{generator/box-plot}
\begin{table}[htbp!]
    \centering
    \caption{Service rates (\%) for each algorithm under various demand levels}
    \label{tab:service_rate}
    \begin{tabular}{|c|c|c|c|c|c|c|c|c|} 
        \hline
        Demand Levels & Full \gls{a:rtv} & Fast \gls{a:rtv} & \gls{a:cg} & \gls{a:la} & \gls{a:la-mr} & \gls{a:la-mr-ns} & \gls{a:la-mr-ps} & \gls{a:la-mr-ce} \\ \hline
        Full & - & 53.67 & 55.39 & 54.63 & 54.66 & 55.21 & 54.94 & \textbf{55.67}\\
        Two-thirds & 70.70 & 70.51 & 68.07 & 70.20 & 70.21 & 70.95 & 70.86 & \textbf{71.61}    \\ 
        One-third & \textbf{93.48} & 93.45 & 92.41 & 93.12 & 93.16 & 93.07 & 93.21 & 93.29     \\ \hline
    \end{tabular}
\end{table}

Table 2 reports the percentage of \gls{a:vmt} for each algorithm relative to \gls{a:la}, which serves as the baseline with 100\% \gls{a:vmt}. These are the same simulations as above where service rate was the primary objective. Likely due to \gls{a:la} being the simplest algorithm, nearly all variants achieve a percentage below 100\%, except \gls{a:la-mr} at one-third and two-thirds demand levels, indicating greater system efficiency. The \gls{a:vmt} results are generally consistent with those of service rates. For instance, \gls{a:la-mr-ce} achieves the lowest \gls{a:vmt} at full and two-thirds demand levels, while the full \gls{a:rtv} achieves the lowest \gls{a:vmt} at the one-third demand level.

It is noticeable that the \gls{a:vmt} reduction achieved by advanced algorithms, relative to \gls{a:la}, exhibits diminishing marginal returns as demand levels increase. This occurs because minimizing \gls{a:vmt} is a secondary objective: with higher demand or when the system can only accommodate a subset of requests, algorithms are more inclined to prioritize assignments and routing decisions that enhance service rates rather than saving \gls{a:vmt}.

\begin{table}[htbp!]
    \centering
    \caption{Percentage of \gls{a:vmt} (\%) for each algorithm with respect to \gls{a:la}}
    \label{tab:vmt}
    \begin{tabular}{|c|c|c|c|c|c|c|c|c|} 
        \hline
        Demand Levels & Full \gls{a:rtv} & Fast \gls{a:rtv} & \gls{a:cg} & \gls{a:la} & \gls{a:la-mr} & \gls{a:la-mr-ns} & \gls{a:la-mr-ps} & \gls{a:la-mr-ce} \\ \hline
        Full & - & 99.44 & 101.00 & 100 & 99.99 & 99.53 & 99.57 & \textbf{99.39}    \\
        Two-thirds & 99.39 & 99.39 & 101.48 & 100 & 100.01 & 99.51 & 99.51 & \textbf{99.33}     \\ 
        One-third & \textbf{96.38} & 96.63 & 110.64 & 100 & 100.04 & 97.78 & 97.56 & 96.73    \\ \hline

    \end{tabular}
\end{table}

Table \ref{tab:computational_time} presents the average full-day computation time for all algorithms. To ensure a fair comparison, we restrict all experiments to a single-thread environment when obtaining their computation times, as different algorithms support parallel computing to varying degrees. One exception is the full \gls{a:rtv} algorithm because it fails to complete the simulation at all three demand levels within $100$ hours when single threaded. The column ``Full \gls{a:rtv} (M)'' presents the results obtained with multi-threading.

Even though multi-threading is enabled, the computational time for the full \gls{a:rtv} algorithm increases drastically as the system scales, far exceeding other methods and rendering it impractical for real-world instances. As heuristic methods, fast \gls{a:rtv} and \gls{a:cg} are faster than full \gls{a:rtv}, but they are still one order of magnitude slower than other algorithms, especially when the demand level is low for \gls{a:cg} and when the demand level is high for fast \gls{a:rtv}.

\gls{a:la} is the most time-efficient algorithm as expected since it is the lowest complexity scheme discussed in this paper. Built on top of \gls{a:la}, all newly introduced ride-pool assignment algorithms require various amounts of extra time to execute the heuristics, including the time for sampling, extending the bipartite graph, solving more sub-problems, and so on. \gls{a:la-mr} performs similarly to \gls{a:la} in terms of service rate but costs roughly 1.5x computation time compared to \gls{a:la}, leading us to conclude that \gls{a:la-mr} would not be a preferred algorithm standing on its own despite it being a valuable skeleton for building the swapping-based algorithms. \gls{a:la-mr-ns} and \gls{a:la-mr-ps} strike a balance between the solution quality and computational efficiency. \gls{a:la-mr-ce}, though being the most costly algorithm among all swapping-based counterparts, shows superiority given that it is considerably faster than the full \gls{a:rtv} algorithm while achieving nearly the same or better service rate.
\begin{table}[htbp!]
    \centering
    \caption{Average full-day simulation time in minutes for each algorithm}
    \label{tab:computational_time}
    \begin{tabular}{|c|c|c|c|c|c|c|c|c|} 
        \hline
        Demand Levels & Full \gls{a:rtv} (M)\tablefootnote{These experiments are executed in a multi-threaded setting with a maximum of 10 concurrent threads.} & Fast \gls{a:rtv} & \gls{a:cg} & \gls{a:la} & \gls{a:la-mr} & \gls{a:la-mr-ns} & \gls{a:la-mr-ps} & \gls{a:la-mr-ce} \\ \hline
        Full & - & 1087.2 & 1922.5 & 19.8 & 29.8 & 160.0 & 142.7 & 502.9 \\
        Two-thirds & 3926.4 & 514.7 & 1134.7 & 13.4 & 19.1 & 111.0 & 93.0 & 334.0     \\
        One-third & 357.6 & 147.5 & 602.7 & 9.3 & 12.0 & 52.7 & 52.5 & 156.4    \\ \hline
    \end{tabular}
\end{table}

Despite the performance improvements and trade-offs introduced by the novel swapping heuristics, the variation in service rates across all algorithms remains minimal, with differences of less than two percent, except for \gls{a:cg}. This similarity in performance raises the question of why different algorithms yield comparable results, motivating a deeper analysis of the epoch-wise solution correlation across successive epochs in Section \ref{sec:correlation}.

\subsection{Correlation of Solutions across Epochs} \label{sec:correlation}
In this section, we further analyze why the aforementioned algorithms perform similarly in the full-day simulation, despite having a larger variance in their potential to solve an epoch-wise \gls{a:rap}. We hypothesize that this similarity is due to the natural feedback mechanisms inherent in the dynamical system. For instance, if an assignment method successfully assigns a large number of requests in one iteration, the subsequent iteration will likely face greater difficulty assigning new requests due to the reduced available system capacity. Conversely, if a method performs poorly in one iteration, the system will have more capacity available in future iterations, allowing it a chance to catch up.

To test this hypothesis, we compare two algorithms, \gls{a:la} and full \gls{a:rtv}, which are at opposite ends of the epoch-wise performance spectrum, under the condition of a uniform demand level throughout the day. The demand is set to 180 requests per minute, and the O-D pair of each request is also uniformly distributed within the boundary. The aim of this trial is to determine whether strong performance by an assignment algorithm (i.e., assigning many requests) is subsequently counteracted by poorer performance in following iterations. With uniform demand levels, we expect the number of assignments per iteration to show no time-based correlation if our hypothesis is false. However, if the performance of one iteration does impact future iterations, we would expect to observe a negative correlation in the number of assignments over time. This would indicate that good performance in one iteration is followed by poor performance, and vice versa.

\begin{figure}[htbp!]
    \centering
    \begin{subfigure}[b]{0.49\textwidth}
      \includegraphics[width=\textwidth]{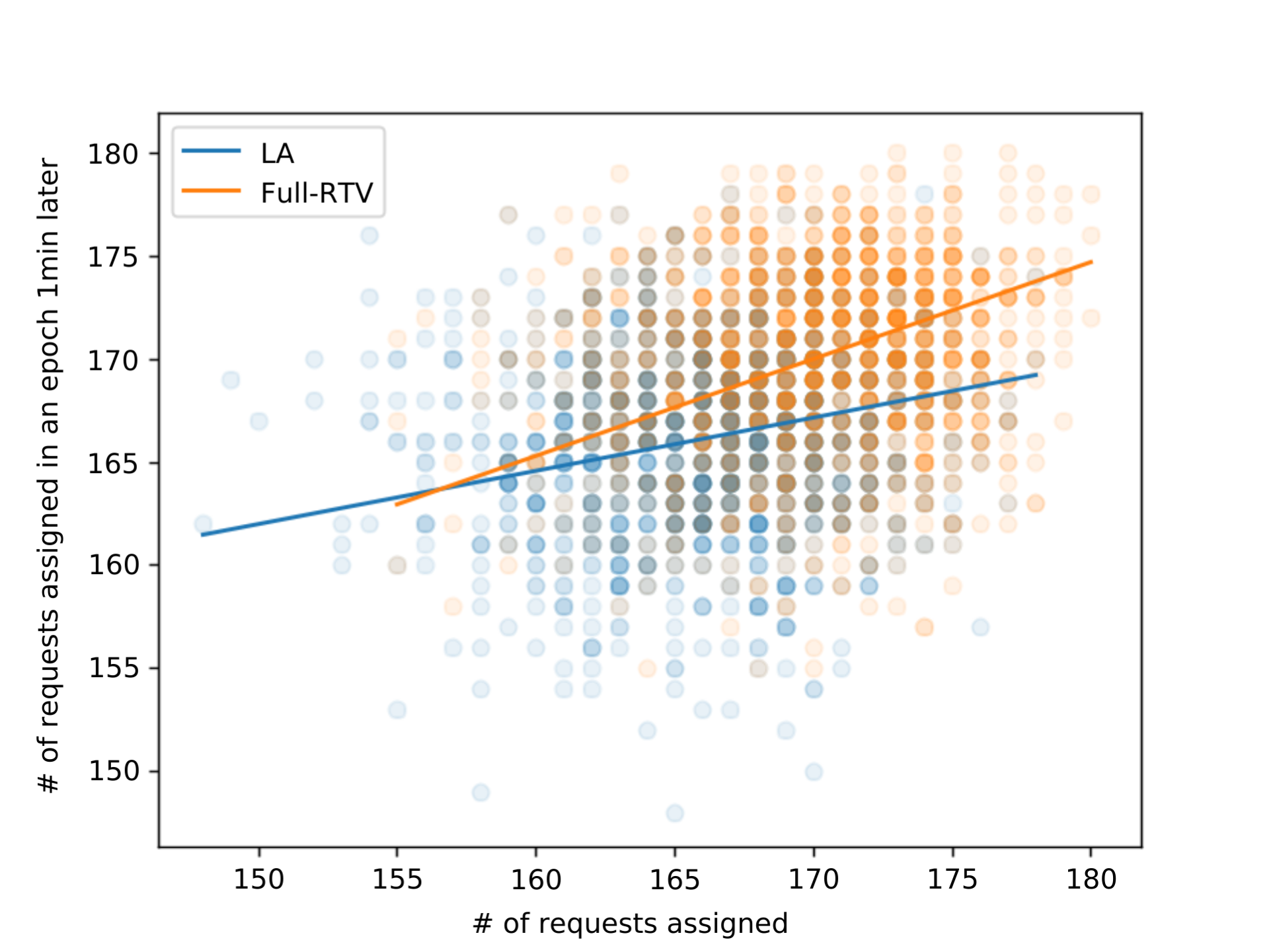}
      \caption{1 minute Lag}
    \end{subfigure}
    \begin{subfigure}[b]{0.49\textwidth}
      \includegraphics[width=\textwidth]{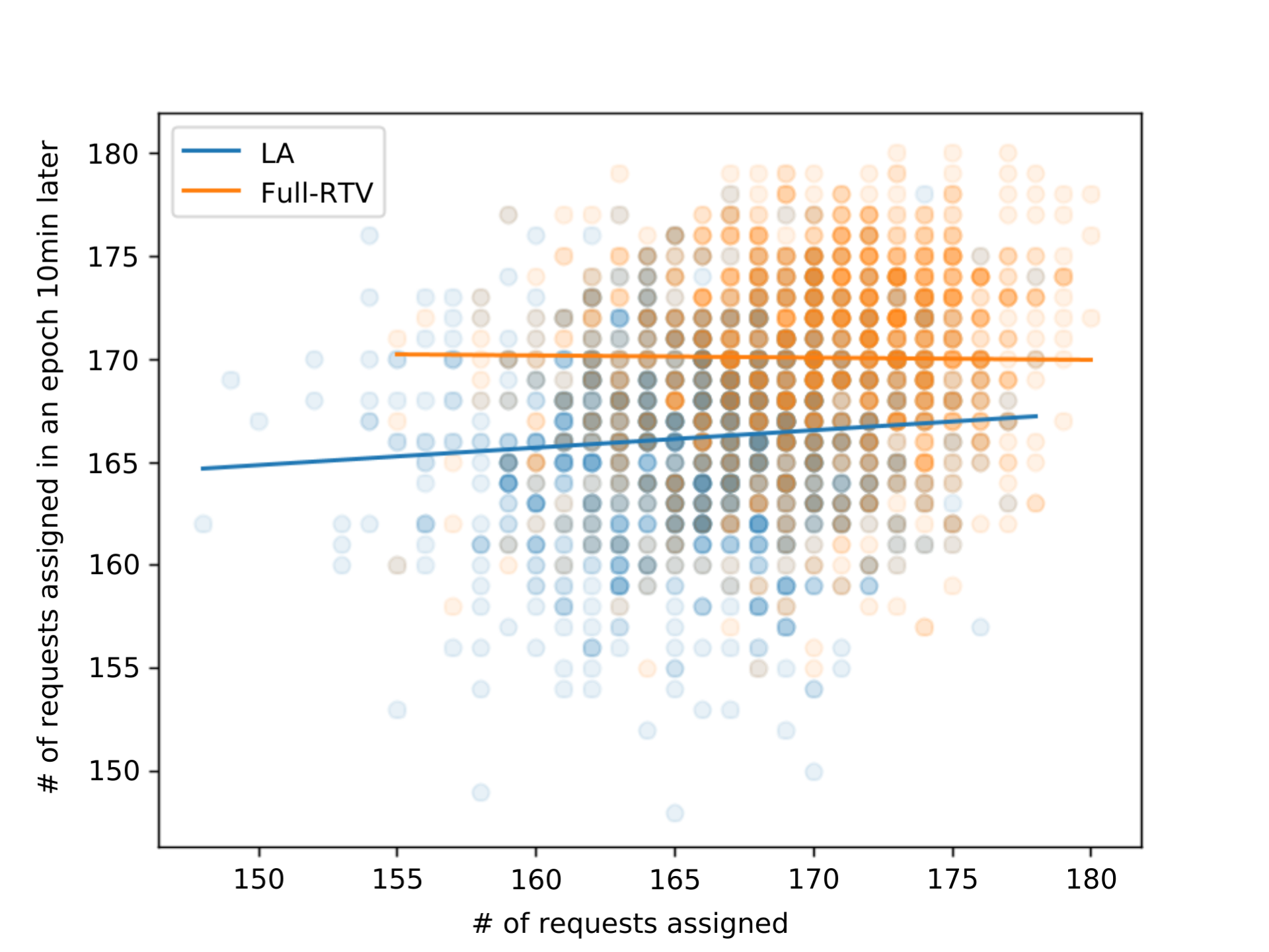}
      \caption{10 minute Lag}
    \end{subfigure}
    
    \begin{subfigure}[b]{0.49\textwidth}
      \includegraphics[width=\textwidth]{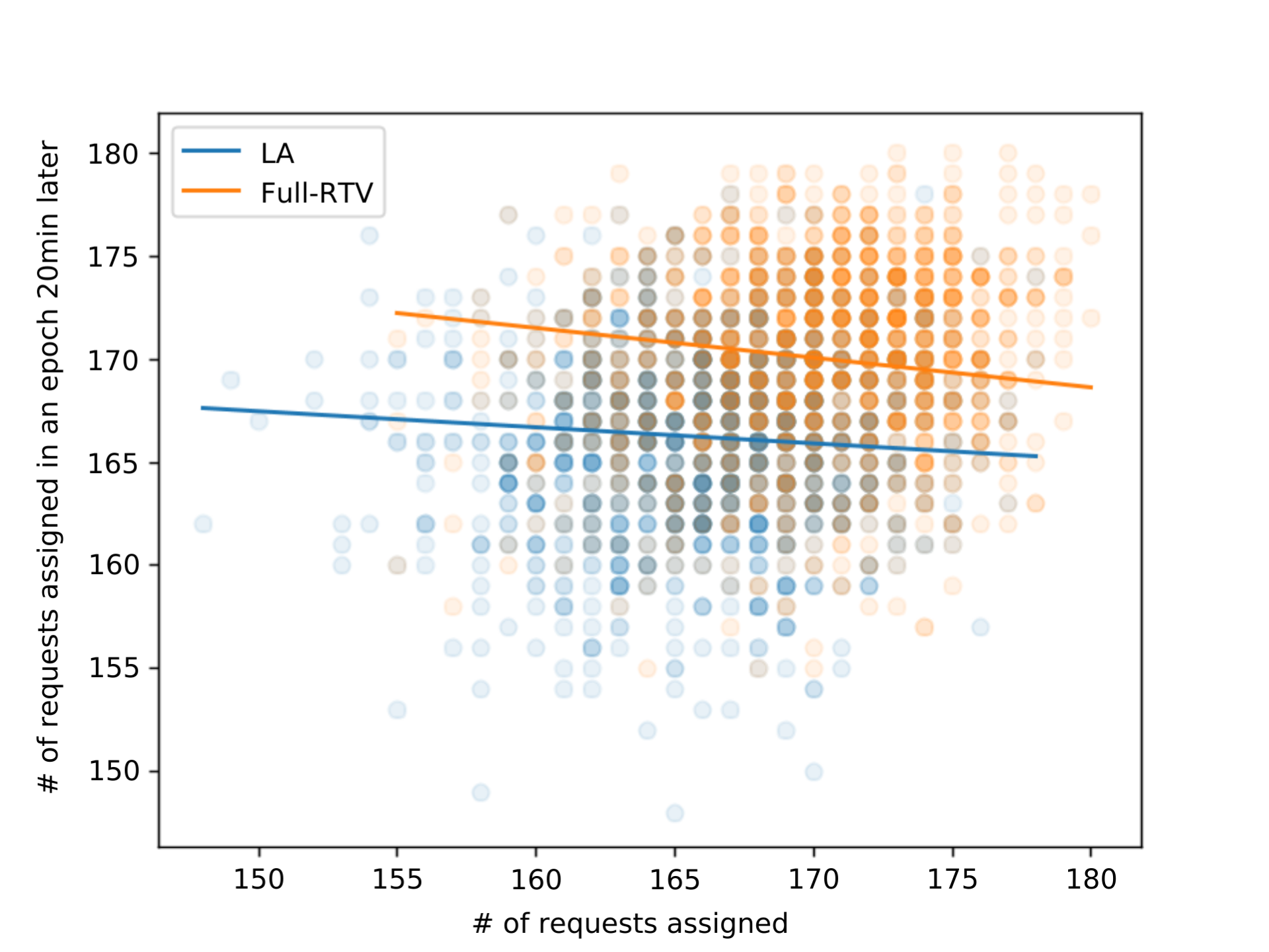}
      \caption{20 minute Lag}
    \end{subfigure}
    \begin{subfigure}[b]{0.49\textwidth}
      \includegraphics[width=\textwidth]{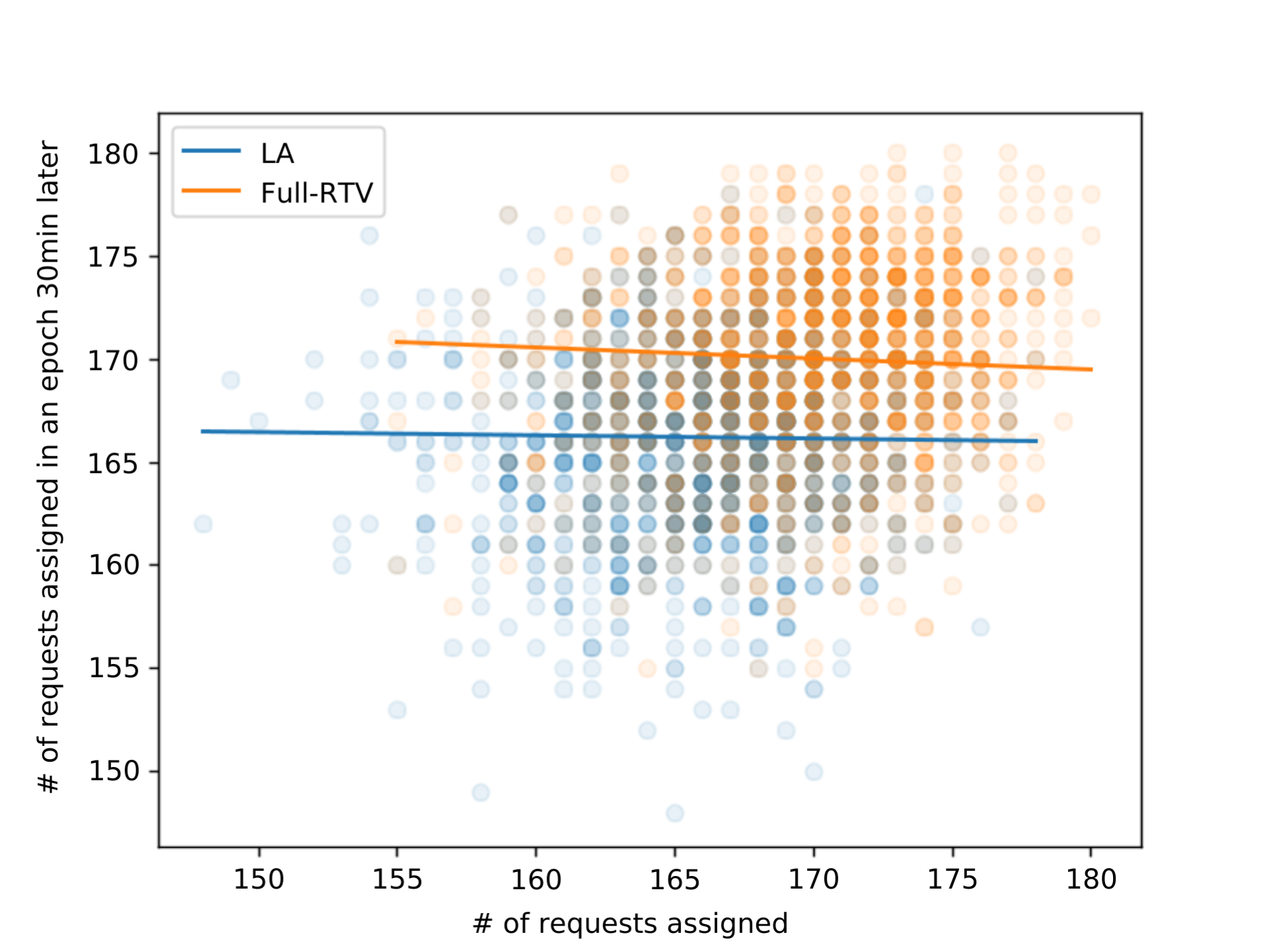}
      \caption{30 minute Lag}
    \end{subfigure}
    
    \caption{Correlation between numbers of served requests in current and future epochs}
    \label{fig:lag_group}
\end{figure}

In Figure \ref{fig:lag_group}, we illustrate how the number of assigned requests changes over time. The scatter plots depict pairs representing the number at the current epoch and the epochs lagging by 1, 10, 20, and 30 minutes. Blue points correspond to \gls{a:la} while orange points correspond to full \gls{a:rtv}. Additionally. regression lines are provided to indicate the trends.

In Figure \ref{fig:lag_group}(a), we observe a weakly positive correlation for the 1-minute lag (i.e., the slopes of regression lines are less than $1$). This suggests that the number of requests assigned at each epoch is somewhat consistent with those assigned in the immediately subsequent epochs. In other words, epochs where larger (smaller) numbers of requests are assigned tend to be followed by epochs that also have a larger (smaller) number of requests assigned. This observation makes sense because the cumulative demand over a short period does not exhaust the system's capacity. Consequently, this would suggest full \gls{a:rtv} retains its advantage over \gls{a:la} in the near-future epochs.


However, in the case of $20$-minute lag shown in Figure \ref{fig:lag_group}(c), we observe a weakly negative correlation. This indicates that if many requests are assigned right now, fewer will be assigned in an epoch 20 minutes later, and vice versa. This suggests that a high-quality solution to the current epoch-wise \gls{a:rap} leads to system states where fewer requests can be assigned. This impact diminishes when the time lag is sufficiently large, as demonstrated by the low correlations shown in Figure \ref{fig:lag_group}(d). Interestingly, the negative impact is more pronounced in full \gls{a:rtv} compared to \gls{a:la}. This is likely due to the greater exploitation of the system's capacity by full \gls{a:rtv} in previous epochs. This would seem to verify our hypothesis that an algorithm with stronger performance tends to be counteracted by system dynamics over time.

\begin{figure}[htbp!]
    \centering
    \begin{minipage}{0.49\textwidth}
    \centering
    \includegraphics[scale=0.5]{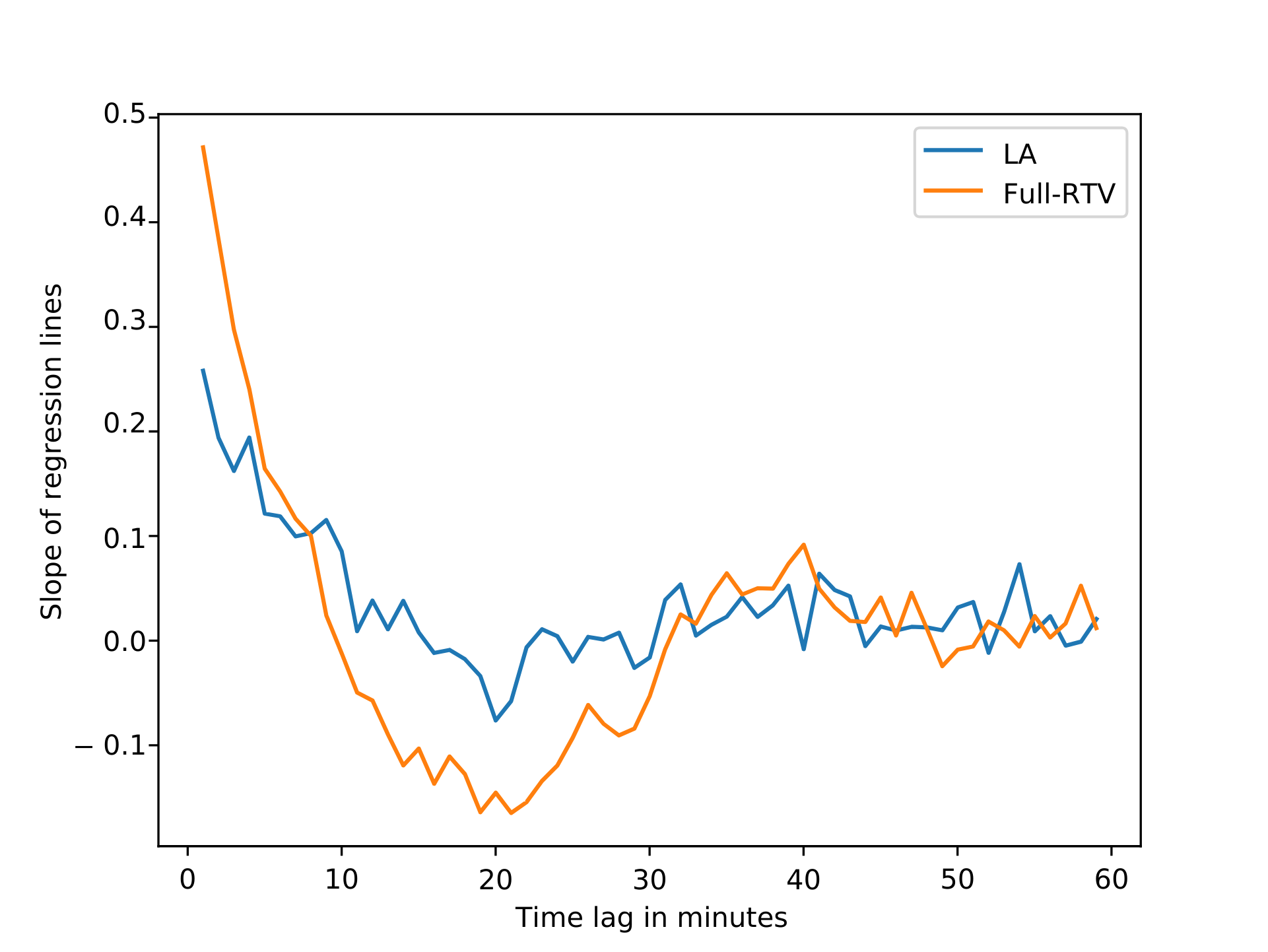}
    \caption{Slopes of the regression lines over time}
    \label{fig:slope_lag}
    \end{minipage}
    \begin{minipage}{0.49\textwidth}
    \centering
    \includegraphics[scale=0.5]{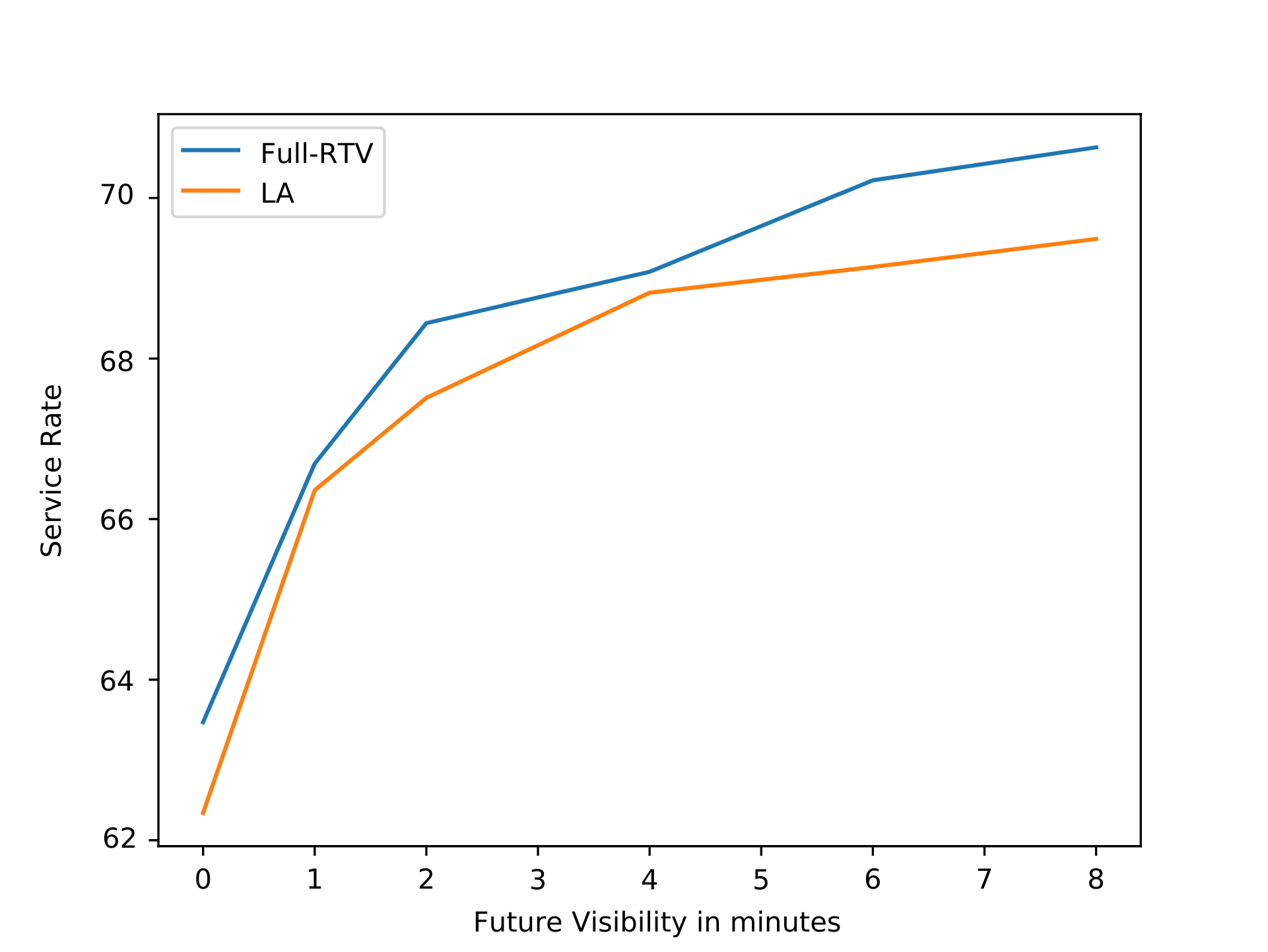}
    \caption{Service rate with future visibility}
    \label{fig:future_visibility}
    \end{minipage}
\end{figure}

To illustrate the analysis above more directly, the slopes of the regression lines are plotted against time lags in Figure \ref{fig:slope_lag}. The results show that the full \gls{a:rtv} algorithm exhibits a negative correlation within a time window of approximately 20 minutes, from $t = 10$ min to $t = 30$ min, whereas \gls{a:la} has a shorter time window and smaller magnitude curve. In conclusion, the system's capacity is packed and emptied over cycles. Algorithms with greater potential in solving the epoch-wise \gls{a:rap} have a larger momentum and thus are subject to larger oscillations. This is the core reason any algorithm capable of assigning requests at a higher rate per epoch than the system's ``capacity'' is limited by the capacity bottleneck and eventually reaches similar performance in the full-day service rate. In fact, we hypothesize this barrier is inherent for all myopic ride-pool assignment algorithms and we hypothesize that integrating extra information from future demand may be the only way to overcome this barrier. 



\subsection{Value of Future Information} \label{sec:future}
In this section, we explore the value of exploiting the future information in the system if any of it is available in any form. It is perhaps more important to directly address the question of how much we stand to gain from knowing the future. Or equivalently, what is the cost of being myopic in the ride-pool assignment problem? While there are many ways one could think of incorporating future information, to get an initial sense, we quantitatively analyze what happens when we have partial visibility into the future.

In this experiment, we use the same set of requests in Section \ref{sec:full_day} but further sample $1/16$ out of it for tractability due to the complexity added by appending future requests to trips. The number of vehicles is reduced to $125$. We use a variant of the \gls{a:lrp} heuristic described in Section \ref{sec:stability}, whereby we allow an excess number of requests to be considered if they are from the future. We first build a solution using the limit and recall prefix \gls{a:ctsp} heuristic without future requests. Then, the future requests are added sequentially to the route, ordered in a way that requests arriving sooner are considered first. While this heuristic variant might not be stable, it can be mitigated by removing the requirement that an assigned future request must also be assigned in subsequent decision epochs.

We consider future request visibility ranging from $0$ up to $8$ minutes. The results, shown in Figure \ref{fig:future_visibility} for both the \gls{a:la} and full \gls{a:rtv} algorithms, demonstrate a substantial increase in service rate with even a small amount of future visibility: $8$ minutes of future visibility raises the service rate from 63.5\% to 70.6\% by full \gls{a:rtv} and from 62.3\% to 69.5\% by \gls{a:la}. A possible explanation for the observed increase in performance may have to do with the coverage. If problem constraints require all requests to be picked up within $5$ minutes of entering the system, then each vehicle acts as a $5$-minute radius cover of the map. When we can see $8$ minutes into the future, the radius of coverage expands to $13$ minutes. While still limited by the ability to pack passengers onto vehicles, this allows planning algorithms to break the information barrier and find better routes that utilize the expanded coverage.

\section{Conclusions} \label{sec:conclusion}
In this work, we present the implementation of a ride-pooling simulator, while thoroughly examining various ride-pool assignment algorithms and essential components, such as the vehicle routing oracle. The detailed technical discussions, combined with the open-source nature of the codebase, equip researchers and practitioners with the tools needed to simulate real-world ride-pooling systems at scale, using their own datasets and objectives. The C++ codebase is designed to be modular and extendable, facilitating the seamless integration of new features and algorithms as needed.

Among the algorithms explored, those based on swapping heuristics stand out as novel contributions to the ride-pool assignment literature. Our extensive experiments indicate that this local search strategy — by swapping assigned requests among available vehicles — achieves a better balance between performance and computational efficiency.

Moreover, we provide evidence of a throughput bottleneck in request assignment within ride-pooling systems. This is demonstrated via the cycling behavior observed in the full \gls{a:rtv} and \gls{a:la} algorithms on a synthetic dataset with uniformly distributed demand. Such a phenomenon explains the similar service rates we observed across all algorithms when applied to the NYC taxi dataset.

Looking forward, we believe that the next advancement in ride-pool assignment lies in the incorporation of future information. As demonstrated in Section \ref{sec:future}, allowing both the full \gls{a:rtv} and \gls{a:la} algorithms to anticipate future demand by eight minutes leads to substantial improvements in service rates. Notably, both algorithms experience similar gains, suggesting that the limitations of myopic optimization may now be more consequential than the differences between the methods used to solve the epoch-wise assignment problem itself.

\newpage

\bibliographystyle{acm}
\bibliography{refs.bib}

\end{document}

%% file: macro.tex
\usepackage[acronym]{glossaries} 
\usepackage{physics,amsmath}

\newacronym{a:rap}{RAP}{Ridepool Assignment Problem}
\newacronym{a:qos}{QoS}{Quality-of-Service}
\newacronym{a:ctsp}{CTSP}{Constrained Traveling Salesman Problem}
\newacronym{a:tsp}{TSP}{Traveling Salesman Problem}
\newacronym{a:rtv}{RTV}{Request-Trip-Vehicle}
\newacronym{a:vmt-minute}{VMT}{Vehicle Minutes Traveled}
\newacronym{a:vmt}{VMT}{Vehicle Miles Traveled}
\newacronym{a:sr}{SR}{Service Rate}
\newacronym{a:rmt}{RMT}{Revenue Miles Traveled}
\newacronym{a:rl}{RL}{Reinforcement Learning}
\newacronym{a:ilp}{ILP}{Integer Linear Program}
\newacronym{a:mip}{MIP}{Mixed Integer Programming}
\newacronym{a:lp}{LP}{Linear Programming}
\newacronym{a:cg}{CG}{Column Generation}
\newacronym{a:la}{LA}{Linear Assignment}
\newacronym{a:rmp}{RMP}{Restricted Master Problem}
\newacronym{a:pp}{PP}{Pricing Problem}
\newacronym{a:la-mr}{LA-MR}{Multi-Round Linear Assignment}
\newacronym{a:la-mr-ns}{LA-MR-NS}{Multi-Round Linear Assignment with Naive Swapping}
\newacronym{a:la-mr-ps}{LA-MR-PS}{Multi-Round Linear Assignment with Proper Swapping}
\newacronym{a:la-mr-ce}{LA-MR-CE}{Multi-Round Linear Assignment with Cyclic Exchange}
\newacronym{a:oof}{OOF}{On-board Order Fixing}
\newacronym{a:lrp}{LRP}{Limit and Recall Prefix}

%% file: generator/la-mr.tex
\begin{figure}[htbp!]
    \centering

    \begin{subfigure}[b]{0.33\textwidth}
    \centering
        \begin{tikzpicture}
        
          \tikzstyle{R}=[circle, draw, minimum size=15pt, inner sep=0pt]
          \tikzstyle{V}=[rectangle, draw, minimum size=15pt, inner sep=0pt]

            \node[text=black, font=\small] at (0,1.75) {Requests};
            \node[text=black, font=\small] at (2,1.75) {Vehicles};
        
          \foreach \i in {1,2,3,4,5}
            \node[R] (A\i) at (0, 2-\i) {\i};
        
          \foreach \i [count=\n from 1] in {1,2,3}
            \node[V] (B\i) at (2, 1-\i) {\i};
        
            \draw[line width=1pt, opacity=0.5, dashed] (A1) -- (B1);
            \draw[line width=1pt] (A2) -- (B1);
            \draw[line width=1pt, opacity=0.5, dashed] (A3) -- (B1);
            
            \draw[line width=1pt] (A3) -- (B2); 
            
            \draw[line width=1pt] (A4) -- (B3);
            \draw[line width=1pt, opacity=0.5, dashed] (A5) -- (B3);

            \draw[line width=4pt, color=red, opacity=0.3] (A2) -- (B1);
            \draw[line width=4pt, color=red, opacity=0.3] (A4) -- (B3);

            \node[right, text=black, font=\small] at (2.3, 0) {\{2\}};
            \node[right, text=black, font=\small] at (2.3, -2) {\{4\}};
            
        \end{tikzpicture}
        \caption{Round 1}
    \end{subfigure}
    \hfill
    \begin{subfigure}[b]{0.33\textwidth}
    \centering
        \begin{tikzpicture}
          \tikzstyle{R}=[circle, draw, minimum size=15pt, inner sep=0pt]
          \tikzstyle{V}=[rectangle, draw, minimum size=15pt, inner sep=0pt]

            \node[text=black, font=\small] at (0,1.75) {Requests};
            \node[text=black, font=\small] at (2,1.75) {Vehicles};
        
          \foreach \i in {1,3,5}
            \node[R] (A\i) at (0, 2-\i) {\i};
          \foreach \i in {2,4}
            \node[R] (A\i)[fill=gray!40] at (0, 2-\i) {\i};
        
          \foreach \i [count=\n from 1] in {1,2,3}
            \node[V] (B\i) at (2, 1-\i) {\i};
        
            \draw[line width=1pt] (A1) -- (B1);
          
            \draw[line width=1pt, opacity=0.5, dashed] (A3) -- (B1); 
            \draw[line width=1pt] (A3) -- (B2); 
            
            \draw[line width=1pt] (A5) -- (B3);

            \draw[line width=4pt, color=red, opacity=0.3] (A1) -- (B1);
            \draw[line width=4pt, color=red, opacity=0.3] (A5) -- (B3);
            
            \node[right, text=black, font=\small] at (2.3, 0) {\{2, 1\}};
            \node[right, text=black, font=\small] at (2.3, -2) {\{4, 5\}};
            
        \end{tikzpicture}
        \caption{Round 2}
    \end{subfigure}
    \hfill
    \begin{subfigure}[b]{0.33\textwidth}
    \centering
        \begin{tikzpicture}
          \tikzstyle{R}=[circle, draw, minimum size=15pt, inner sep=0pt]
          \tikzstyle{V}=[rectangle, draw, minimum size=15pt, inner sep=0pt]

            \node[text=black, font=\small] at (0,1.75) {Requests};
            \node[text=black, font=\small] at (2,1.75) {Vehicles};
        
          \foreach \i in {1,2,4,5}
            \node[R] (A\i)[fill=gray!40] at (0, 2-\i) {\i};
          
          \node[R] (A3) at (0, -1) {3};
        
          \foreach \i [count=\n from 1] in {1,2,3}
            \node[V] (B\i) at (2, 1-\i) {\i};
        
          \draw[line width=1pt] (A3) -- (B1); 
          \draw[line width=1pt, opacity=0.5, dashed] (A3) -- (B2); 

            \draw[line width=4pt, color=red, opacity=0.3] (A3) -- (B1);

            \node[right, text=black, font=\small] at (2.3, 0) {\{2, 1, 3\}};
            \node[right, text=black, font=\small] at (2.3, -2) {\{4, 5\}};

        \end{tikzpicture}
        \caption{Round 3}
    \end{subfigure}
    
    \caption{A schematic diagram illustrating the \gls{a:la-mr} algorithm}
    \label{fig:la-mr}
\end{figure}

%% file: generator/la-mr-ns.tex
\begin{figure}[htbp!]
    \centering

    \begin{subfigure}[b]{0.33\textwidth}
    \centering
        \begin{tikzpicture}
        
          \tikzstyle{R}=[circle, draw, minimum size=15pt, inner sep=0pt]
          \tikzstyle{V}=[rectangle, draw, minimum size=15pt, inner sep=0pt]

            \node[text=black, font=\small] at (0,1.75) {Requests};
            \node[text=black, font=\small] at (2,1.75) {Vehicles};
        
          \foreach \i in {1,2,3,4}
            \node[R] (A\i) at (0, 2-\i) {\i};
        
          \foreach \i [count=\n from 1] in {1,2,3}
            \node[V] (B\i) at (2, 1-\i) {\i};
        
            \draw[line width=1pt, opacity=0.5, dashed] (A1) -- (B1);
            \draw[line width=1pt] (A2) -- (B1);
            \draw[line width=1pt, opacity=0.5, dashed] (A3) -- (B1);

            \draw[line width=1pt, opacity=0.5, dashed] (A2) -- (B2);
            \draw[line width=1pt] (A3) -- (B2);
            \draw[line width=1pt, opacity=0.5, dashed] (A4) -- (B2);

            \draw[line width=1pt, opacity=0.5, dashed] (A3) -- (B3);
            \draw[line width=1pt] (A4) -- (B3);

            \draw[line width=4pt, color=red, opacity=0.3] (A2) -- (B1);
            \draw[line width=4pt, color=red, opacity=0.3] (A3) -- (B2);
            \draw[line width=4pt, color=red, opacity=0.3] (A4) -- (B3);   

            \node[right, text=black, font=\small] at (2.3, 0) {\{2\}};
            \node[right, text=black, font=\small] at (2.3, -1) {\{3\}};
            \node[right, text=black, font=\small] at (2.3, -2) {\{4\}};
            
        \end{tikzpicture}
        \caption{Round 1}
    \end{subfigure}
    \hfill
    \begin{subfigure}[b]{0.33\textwidth}
    \centering
        \begin{tikzpicture}
          \tikzstyle{R}=[circle, draw, minimum size=15pt, inner sep=0pt]
          \tikzstyle{V}=[rectangle, draw, minimum size=15pt, inner sep=0pt]
          \tikzstyle{S}=[diamond, draw, minimum size=15pt, inner sep=0pt]

            \node[text=black, font=\small] at (0,1.75) {Requests};
            \node[text=black, font=\small] at (2,1.75) {Vehicles};
        
          \node[R] (A1) at (0, 1) {1};

          \foreach \i in {2,3,4}{
            \node[S] (C\i) at (0, 2-\i) {\i};
            \node[R] (A\i)[fill=gray!40] at (-0.75, 2-\i) {\i};
            }
        
          \foreach \i [count=\n from 1] in {1,2,3}
            \node[V] (B\i) at (2, 1-\i) {\i};
        
            \draw[line width=1pt] (A1) -- (B1);
            \draw[line width=1pt, opacity=0.5, dashed] (C3) -- (B1);

            \draw[line width=1pt] (C2) -- (B2);
            \draw[line width=1pt, opacity=0.5, dashed] (C4) -- (B2);

            \draw[line width=1pt] (C3) -- (B3);

            \draw[line width=4pt, color=red, opacity=0.3] (A1) -- (B1);
            \draw[line width=4pt, color=red, opacity=0.3] (C3) -- (B3);   

            \node[right, text=black, font=\small] at (2.3, 0) {\{2, 1\}};
            \node[right, text=black, font=\small] at (2.3, -2) {\{4, 3\}};
            
        \end{tikzpicture}
        \caption{Round 2}
    \end{subfigure}
    \hfill
    \begin{subfigure}[b]{0.33\textwidth}
    \centering
        \begin{tikzpicture}
          \tikzstyle{R}=[circle, draw, minimum size=15pt, inner sep=0pt]
          \tikzstyle{V}=[rectangle, draw, minimum size=15pt, inner sep=0pt]
          \tikzstyle{S}=[diamond, draw, minimum size=15pt, inner sep=0pt]

            \node[text=black, font=\small] at (0,1.75) {Requests};
            \node[text=black, font=\small] at (2,1.75) {Vehicles};

          \foreach \i in {1,2,3,4}{
            \node[R] (A\i)[fill=gray!40] at (-0.75, 2-\i) {\i};
            }
          \foreach \i in {1,4}
            \node[S] (C\i)[fill=gray!40] at (0, 2-\i) {\i};
          \foreach \i in {2,3}
            \node[S] (C\i) at (0, 2-\i) {\i};
        
          \foreach \i [count=\n from 1] in {1,2,3}
            \node[V] (B\i) at (2, 1-\i) {\i};
        
            \draw[line width=1pt] (C3) -- (B1);
            \draw[line width=1pt] (C2) -- (B2);

            \draw[line width=4pt, color=red, opacity=0.3] (C2) -- (B2);   

            \node[right, text=black, font=\small] at (2.3, 0) {\{1\}};
            \node[right, text=black, font=\small] at (2.3, -1) {\{2\}};
            \node[right, text=black, font=\small] at (2.3, -2) {\{4, 3\}};


        \end{tikzpicture}
        \caption{Round 3}
    \end{subfigure}
    
    \caption{A schematic diagram illustrating the \gls{a:la-mr-ns} algorithm}
    \label{fig:la-mr-ns}
\end{figure}

%% file: generator/la-mr-ns-2.tex
\begin{figure}[htbp!]
    \centering

    \begin{subfigure}[b]{0.49\textwidth}
    \centering
        \begin{tikzpicture}
        
          \tikzstyle{R}=[diamond, draw, minimum size=17.5pt, inner sep=0pt]
          \tikzstyle{V}=[rectangle, draw, minimum size=15pt, inner sep=0pt]

            \node[text=black, font=\small] at (0,2.75) {Requests};
            \node[text=black, font=\small] at (2,2.75) {Vehicles};
        
            \node[R] (A1) at (0, 2) {1};
            \node[R] (A2) at (0, 0) {2};
        
            \node[V] (B1) at (2, 2) {1};
            \node[V] (B2) at (2, 0) {2};
        
            \draw[line width=1pt, opacity=0.5, dashed] (A1) -- (B1);
            \draw[line width=1pt, opacity=0.5, dashed] (A2) -- (B2);
            \draw[line width=1pt] (A2) -- (B1) node[pos=0.675, right] {2};
            \draw[line width=1pt] (A1) -- (B2) node[pos=0.4, left] {2};

            \node[right, text=black, font=\small] at (2.5, 2.2) {before:};
            \node[right, text=black, font=\small] at (2.5, 1.8) {after:};
            \node[right, text=black, font=\small] at (3.5, 2.2) {\{1\}};
            \node[right, text=black, font=\small] at (3.5, 1.8) {\{2\}};

            \node[right, text=black, font=\small] at (2.5, 0.2) {before:};
            \node[right, text=black, font=\small] at (2.5, -0.2) {after:};
            \node[right, text=black, font=\small] at (3.5, 0.2) {\{2\}};
            \node[right, text=black, font=\small] at (3.5, -0.2) {\{1\}};
            
        \end{tikzpicture}
        \caption{The bipartite graph of two vehicles and two requests}
    \end{subfigure}
    \hfill
    \begin{subfigure}[b]{0.49\textwidth}
    \centering
        \begin{tikzpicture}
          \tikzstyle{R}=[circle, draw, minimum size=15pt, inner sep=0pt]
          \tikzstyle{V}=[rectangle, draw, minimum size=15pt, inner sep=0pt]

          \node[R] (A1) at (0, 0) {1o};
          \node[R] (A2) at (0, -1) {1d};
          \node[R] (B1) at (1, 0) {2o};
          \node[R] (B2) at (1, -1) {2d};
          
          \node[V] (C1) at (-2, 2) {1};
          \node[V] (C2) at (3, 2) {2};

            \draw[line width=1pt] (C1) -- (A1) node[midway, above] {3};
            \draw[line width=1pt] (C2) -- (B1) node[midway, above] {3};
            \draw[line width=1pt] (A1) -- (A2) node[midway, left] {1};
            \draw[line width=1pt] (A1) -- (B1) node[midway, above] {1};
            \draw[line width=1pt] (B1) -- (B2) node[midway, right] {1};
            \draw[line width=1pt] (A2) -- (B2) node[midway, above] {1};            
            
        \end{tikzpicture}
        \caption{The graph in Euclidean space showing the locations}
    \end{subfigure}
    
    \caption{An example illustrating the defect of naive swapping on dependent vehicles. In Figure (a), dashed edges represent the current assignment while solid edges represent the valid naive swaps. The numbers next to the solid edges indicate cost reduction. In Figure (b), the same problem is shown where squares are vehicles while circles denote request origin-destination pairs. The numbers next to the edges indicate the distances. Although both naive swaps reduce costs on their own, applying both simultaneously leads to a worse assignment.}
    \label{fig:la-mr-ns-2}
\end{figure}

%% file: generator/la-mr-ps.tex
\begin{figure}[htbp!]
    \centering

    \begin{subfigure}[b]{0.33\textwidth}
    \centering
        \begin{tikzpicture}
        
          \tikzstyle{R}=[circle, draw, minimum size=15pt, inner sep=0pt]
          \tikzstyle{V}=[rectangle, draw, minimum size=15pt, inner sep=0pt]

            \node[text=black, font=\small] at (0,1.75) {Requests};
            \node[text=black, font=\small] at (2,1.75) {Vehicles};
        
          \foreach \i in {1,2,3,4}
            \node[R] (A\i) at (0, 2-\i) {\i};
        
          \foreach \i [count=\n from 1] in {1,2,3}
            \node[V] (B\i) at (2, 1-\i) {\i};
        
            \draw[line width=1pt, opacity=0.5, dashed] (A1) -- (B1);
            \draw[line width=1pt] (A2) -- (B1);
            \draw[line width=1pt, opacity=0.5, dashed] (A3) -- (B1);

            \draw[line width=1pt, opacity=0.5, dashed] (A2) -- (B2);
            \draw[line width=1pt] (A3) -- (B2);
            \draw[line width=1pt, opacity=0.5, dashed] (A4) -- (B2);

            \draw[line width=1pt, opacity=0.5, dashed] (A3) -- (B3);
            \draw[line width=1pt] (A4) -- (B3);

            \draw[line width=4pt, color=red, opacity=0.3] (A2) -- (B1);
            \draw[line width=4pt, color=red, opacity=0.3] (A4) -- (B3);   

            \node[right, text=black, font=\small] at (2.3, 0) {\{2\}};
            \node[right, text=black, font=\small] at (2.3, -2) {\{4\}};
            
        \end{tikzpicture}
        \caption{Round 1}
    \end{subfigure}
    \hfill
    \begin{subfigure}[b]{0.33\textwidth}
    \centering
        \begin{tikzpicture}
          \tikzstyle{R}=[circle, draw, minimum size=15pt, inner sep=0pt]
          \tikzstyle{V}=[rectangle, draw, minimum size=15pt, inner sep=0pt]

            \node[text=black, font=\small] at (0,1.75) {Requests};
            \node[text=black, font=\small] at (2,1.75) {Vehicles};
        
          \node[R] (A1) at (0, 1) {1};
          \node[R] (A3) at (0, -1) {3};
          \node[R] (A2)[fill=gray!40] at (0, 0) {2};
          \node[R] (A4)[fill=gray!40] at (0, -2) {4};

          \foreach \i [count=\n from 1] in {1,2,3}
            \node[V] (B\i) at (2, 2.5-1.5*\i) {\i};
        
            \draw[line width=1pt] (A1) -- (B1);
            \draw[line width=1pt, dashed, opacity=0.5] (A3) -- (B1);
            \draw[line width=1pt, dashed, opacity=0.5] (A3) -- (B2);
            \draw[line width=1pt] (A3) -- (B3);
            \draw[line width=1pt, dashed, opacity=0.5, ->] (B1) -- (B2) node[left, midway] {2};

            \draw[line width=4pt, color=red, opacity=0.3] (A1) -- (B1);

            \node[right, text=black, font=\small] at (2.3, 1) {\{1, 2\}};
            \node[right, text=black, font=\small] at (2.3, -2) {\{4\}};
            
        \end{tikzpicture}
        \caption{Round 2}
    \end{subfigure}
    \hfill
    \begin{subfigure}[b]{0.33\textwidth}
    \centering
        \begin{tikzpicture}
        
          \tikzstyle{R}=[circle, draw, minimum size=15pt, inner sep=0pt]
          \tikzstyle{V}=[rectangle, draw, minimum size=15pt, inner sep=0pt]

            \node[text=black, font=\small] at (0,1.75) {Requests};
            \node[text=black, font=\small] at (2,1.75) {Vehicles};

          \foreach \i in {1,2,4}{
            \node[R] (A\i)[fill=gray!40] at (0, 2-\i) {\i};
            }
          \node[R] (A3) at (0, -1) {3};
        
          \foreach \i [count=\n from 1] in {1,2,3}
            \node[V] (B\i) at (2, 2.5-1.5*\i) {\i};
        
            \draw[line width=1pt, ->] (B1) -- (B2) node[left, midway] {2};
            \draw[line width=1pt, dashed, opacity=0.5] (A3) -- (B1);
            \draw[line width=1pt, dashed, opacity=0.5] (A3) -- (B2);
            \draw[line width=1pt] (A3) -- (B3);

            \draw[line width=4pt, color=red, opacity=0.3] (B1) -- (B2);
            \draw[line width=4pt, color=red, opacity=0.3] (A3) -- (B3);

            \node[right, text=black, font=\small] at (2.3, 1) {\{1\}};
            \node[right, text=black, font=\small] at (2.3, -0.5) {\{2\}};
            \node[right, text=black, font=\small] at (2.3, -2) {\{4, 3\}};

        \end{tikzpicture}
        \caption{Round 3}
    \end{subfigure}
    
    \caption{A schematic diagram illustrating the \gls{a:la-mr-ns} algorithm}
    \label{fig:la-mr-ps}
\end{figure}

%% file: generator/ce.tex
\begin{figure}[htbp!]
    \centering

    \begin{tikzpicture}
      \tikzstyle{R}=[circle, draw, minimum size=15pt, inner sep=0pt]
      \definecolor{darkgreen}{rgb}{0.0, 0.5, 0.0}

      \node[R] (A1) at (0, 0) {1};
      \node[R] (A2) at (1.5, 0) {2};
      \node[R] (A3) at (3, 0) {3};
      \node[R] (A4) at (4.5, 0) {4};
      \node[R] (B1) at (0, 1.5) {5};
      \node[R] (B2) at (1.5, 1.5) {6};
      \node[R] (C1) at (3, 1.5) {7};
      \node[R] (C2) at (4.5, 1.5) {8};

      \draw (-0.5, -0.5)[draw = blue, thick] rectangle (5, 0.5);
      \draw (-0.5, 1)[draw = red, thick] rectangle (2, 2);
      \draw (2.5, 1)[draw = darkgreen, thick] rectangle (5, 2);

      \draw[thick, ->] (A3) -- (C1);
      \draw[thick, ->] (C1) -- (B2);
      \draw[thick, ->] (B2) -- (A3);

      \draw[thick, ->] (A1.west) to [out=135, in=-135] (B1.west);
      \draw[thick, ->] (B1.east) to [out=-45, in=45] (A1.east);

        \node[text=black, font=\small] at (0,0.75) {C1};
        \node[text=black, font=\small] at (2.7,0.75) {C2};

    \end{tikzpicture}

    \caption{The pattern of cyclic exchange in the partition problem}
    \label{fig:ce}
\end{figure}

%% file: generator/la-mr-ce.tex
\begin{figure}[htbp!]
    \centering

    \begin{tikzpicture}
      \tikzstyle{R}=[circle, draw, minimum size=15pt, inner sep=0pt]
      \tikzstyle{V}=[rectangle, draw, minimum size=15pt, inner sep=0pt]
      
      \definecolor{darkgreen}{rgb}{0.0, 0.5, 0.0}


      \node[R] (A1) at (0, 0) {$r_1$};
      \node[R] (A2) at (-1, 1.73) {$r_2$};
      \node[R] (A3) at (1, 1.73) {$r_3$}; 

      \node[V] (B3) at (-2, 0) {$v_3$};
      \node[V] (B2) at (2, 0) {$v_2$};
      \node[V] (B1) at (0, 3.46) {$v_1$};

      
      \draw[thick, dashed, opacity=0.5, ->] (A1) -- (A2);
      \draw[thick, color=red, ->, transform canvas={xshift=-1.3mm, yshift=-0.75mm}] (A2) -- (A1);
      
      \draw[thick, color=red, ->, transform canvas={yshift=1.5mm}] (A3) -- (A2);
      \draw[thick, dashed, opacity=0.5, ->] (A2) -- (A3);
      
      \draw[thick, color=blue, ->] (A3) -- (A1);
      \draw[thick, color=red, ->, transform canvas={xshift=+1.3mm, yshift=-0.75mm}] (A1) -- (A3);

      \draw[thick, dashed, opacity=0.5, <->] (B1) -- (A3);

      \draw[thick, color=blue, ->, transform canvas={xshift=1.3mm, yshift=0.75mm}] (B2) -- (A3);
      \draw[thick, dashed, opacity=0.5, ->] (A3) -- (B2);

      \draw[thick, dashed, opacity=0.5, <->] (A2) -- (B1);

      \draw[thick, dashed, opacity=0.5, <->] (B3) -- (A2);

      \draw[thick, dashed, opacity=0.5, <->] (B3) -- (A1);

      \draw[thick, dashed, opacity=0.5, ->] (B2) -- (A1);
      \draw[thick, color=blue, ->, transform canvas={yshift=-1.5mm}] (A1) -- (B2);

      
      \draw[thick, dashed, opacity=0.5, ->] (B1) -- (A1);
      \draw[thick, dashed, opacity=0.5, ->] (B2) -- (A2);
      \draw[thick, dashed, opacity=0.5, ->] (B3) -- (A3);





    \end{tikzpicture}

    \caption{An example graph for cyclic exchange in \gls{a:la-mr-ce}. Suppose $r_i$ is assigned to $v_i$ for $i \in \{1, 2, 3\}$ in the current partitioning. The red cycle $r_1-r_2-r_3$ represents a real cycle while the blue cycle $r_1-v_2-r_3$ only represents a chain.}
    \label{fig:la-mr-ce}
\end{figure}

%% file: generator/box-plot.tex
\begin{figure}[htbp!]
    \centering
    \begin{tikzpicture}	
    \pgfplotstableread[col sep=comma]{data.csv}\csvdata
    	\pgfplotstabletranspose\datatransposed{\csvdata} 
    	\begin{axis}[
    		boxplot/draw direction = y,
    		x axis line style = {opacity=0},
    		axis x line* = bottom,
    		axis y line = left,
    		enlarge y limits,
    		ymajorgrids,
    		xtick = {1, 2, 3, 4, 5, 6, 7},
    		xticklabel style = {align=center, font=\small, rotate=60},
    		xticklabels = {Fast RTV, CG, LA, LA-MR, LA-MR-NS, LA-MR-PS, LA-MR-CE},
    		xtick style = {draw=none}, 
    		ylabel = {Service rate (\%)},
    		ytick = {52, 57}
    	]
    		\foreach \n in {1,...,7} {
    			\addplot+[boxplot, fill, draw=black] table[y index=\n] {\datatransposed};
    		}
    	\end{axis}
    \end{tikzpicture}
    \caption{Service rates for all algorithms under the full demand level from 2013-03-01 to 2013-03-07}
    \label{fig:box-plot}
\end{figure}